\documentclass{ieeeaccess}
\usepackage{cite}
\usepackage{amsmath,amssymb,amsfonts}
\usepackage{algorithmic}
\usepackage{graphicx}
\usepackage{textcomp}
\usepackage{algorithm}
\usepackage{multirow}
\usepackage{stfloats}
\usepackage{psfrag}

\let\emph\textit

\usepackage{soul}
\soulregister\cite7 
\soulregister\citep7 
\soulregister\citet7 
\soulregister\ref7 
\soulregister\pageref7 

\def\BibTeX{{\rm B\kern-.05em{\sc i\kern-.025em b}\kern-.08em
    T\kern-.1667em\lower.7ex\hbox{E}\kern-.125emX}}
    
\begin{document}
\history{Date of publication xxxx 00, 0000, date of current version xxxx 00, 0000.}
\doi{10.1109/ACCESS.2017.DOI}

\title{A Dataset and Benchmark towards Multi-modal Face Anti-spoofing under Surveillance Scenarios}

\author{\uppercase{Xudong Chen}, 
\uppercase{Shugong Xu}, \IEEEmembership{Fellow, IEEE},
\uppercase{Qiaobin Ji}, 
\uppercase{Shan Cao}, \IEEEmembership{Member, IEEE}}

\address {Shanghai Institute for Advanced Communication and Data Science, Shanghai University, Shanghai, 200444, China\\}

\markboth
{X. Chen \headeretal: A Dataset and Benchmark towards Multi-modal Face Anti-spoofing under Surveillance Scenarios}
{X. Chen \headeretal: A Dataset and Benchmark towards Multi-modal Face Anti-spoofing under Surveillance Scenarios}

\corresp{Corresponding author: Shugong Xu (e-mail: shugong@shu.edu.cn)}

\begin{abstract}
Face Anti-spoofing (FAS) is a challenging problem due to complex serving scenarios and diverse face presentation attack patterns. Especially when captured images are low-resolution, blurry, and coming from different domains, the performance of FAS will degrade significantly. The existing multi-modal FAS datasets rarely pay attention to the cross-domain problems under deployment scenarios, which is not conducive to the study of model performance. To solve these problems, we explore the fine-grained differences between multi-modal cameras and construct a cross-domain multi-modal FAS dataset under surveillance scenarios called GREAT-FASD-S. Besides, we propose an {\bf A}ttention based {\bf F}ace {\bf A}nti-spoofing network with {\bf F}eature {\bf A}ugment (AFA) to solve the FAS towards low-quality face images. It consists of the depthwise separable attention module (DAM) and the multi-modal based feature augment module (MFAM). Our model can achieve state-of-the-art performance on the CASIA-SURF dataset and our proposed GREAT-FASD-S dataset.

\end{abstract}


\maketitle

\section{Introduction}
\label{setc:intro}
Face anti-spoofing is a task to determine whether the input face image is real or fake. Nowadays face recognition and verification\cite{bowyer2004face,poola2017artificial} have been widely deployed in security monitoring, financial authorization, and other surveillance scenarios. Due to the easy availability of the target face data, face recognition systems have been the prime target for deception such as presentation attacks. However, the face anti-spoofing system is still targeting on cooperative scenes, where the environment is controllable and images are clear and high-resolution. In cooperative scenes, the spoofing cues are obvious so it is easy to accomplish the face anti-spoofing problem. Limited by imaging elements, such as the lens optical resolution and the CMOS pixel resolution, the resolution of captured images is limited. When the target is far away from the camera, we can only use a few pixels to represent the target, which will result in loss of information. Because of the limitation of shutter speed, images will be blurry when the target moves too fast. Due to the noise caused by low-quality images\cite{borel2019image}, the spoofing cues in surveillance scenarios can hardly be detected. The existing FAS system can not protect the face recognition system from the risk of being deceived in complex scenes. Besides, the existing face anti-spoofing datasets are always captured using one single camera, differences between imaging devices are rarely considered, which is common in deployment scenes. Therefore, the face anti-spoofing (FAS) method towards low-quality images and the dataset targeting on the cross-device domain problem under surveillance scenarios are needed eagerly. They are important for protecting the face recognition system from being deceived\cite{boulkenafet2016face,boulkenafet2016faceletter,liu2018learning,shao2019multi}.

Most existing face anti-spoofing methods focus on detecting 2D plane replaying facial attacks using one single modality (RGB) data. This type of facial attacks is simple and low cost. In the past few years, several single-modal datasets were released, which promoted the academic development in the single-modal face anti-spoofing field. However, the single-modal face anti-spoofing method has inherent defects. Specific modal data have limited usage scenarios. For example, RGB images fail in varying brightness. IR images fail in very high-temperature regions. Depth images fail once the camera is out of the radial detection zone for the object. Some research\cite{yu2020multi,zhang2019dataset} prove that single-modal FAS systems\cite{jourabloo2018face,liu2018learning} have relatively low performance compared to multi-modal FAS systems\cite{zhang2019feathernets,yu2020multi}. To make matters worse, with the development of 3D printing and material science we can use materials that have a similar texture and diffuse reflection coefficient as human skin to reconstruct the 3D mask\cite{gecer2019ganfit} of the target person. Compared with the conventional 2D attack, 3D mask attacks are much more realistic and can hardly be detected only using the single-modal data. 

To improve the performance and robustness of the single-modal face anti-spoofing, two kinds of methods are proposed in the past several years. One kind of methods are sequence-based verification\cite{pan2007eyeblink,patel2016cross,yang2019face}. In sequence-based methods, users are asked to do a series of actions according to the random given prompts, such as blinking, opening the mouth, turning the head to the left or right, \emph{etc.} During the process of interactive verification, the sequence-based methods can prevent attack methods such as print attacks and screen attacks. But replay attacks using video recording can pass if the action instructions are consistent with the given prompts. The defect of such verification methods is that it should take a long time. If some prompts are not successfully executed, the user should complete the verification process again, which is not efficient. Moreover, actions such as blinking are hardly detected in the low-resolution and blurry images.

The other kind of method utilizes multi-modal face data to accomplish the face anti-spoofing task. With the development of technology in sensor and multi-modal imaging, the cost of multi-modal cameras become affordable. In the face recognition field, researchers\cite{li2013using}\cite{li2016face} use RGB-D data captured by Kinect to train and deploy their models. Compared to face recognition methods only using RGB images or depth images, using multi-modal data can boost performance significantly. Recently researchers\cite{zhang2019dataset,parkin2019recognizing,shen2019facebagnet} pay more attention to the multi-modal FAS. Compared to ordinary RGB cameras, multi-modal cameras can simultaneously capture near-infrared (IR) images, depth images, and RGB images. The depth images and IR images can make up the shortcomings of RGB images. The depth map of a real face is very different from the depth map of common types of plane attacks. So plane attacks can be easily defended in the depth mode\cite{atoum2017face}. Near-infrared (IR) images can capture the material characteristics of the target. The difference in the texture of different materials can be used as a strong basis for judging whether it is a real person\cite{sun2016context,jiang2019multilevel}. Because infrared light is self-luminous and does not depend on environment light, infrared images can be used in low-light or even no-light scenarios. In most cases, comprehensive using the information of these three modalities directly can complete the face anti-spoofing. However, existing multi-modal face anti-spoofing methods are not robust enough to cope with the migration across different device domains, such as data captured using cameras with different imaging principles. Although they can achieve good performance on the test set which is in the same domain as the training data, their performance will decrease significantly after changing to another domain dataset. Retraining the previous model or fine-tuning this model on extra target domain data will be indispensable. Besides, existing multi-modal face anti-spoofing systems do not consider the surveillance scenarios. The performance of these methods will drop significantly when the input images are low-resolution and blurry.

The existing large-scale multi-modal face anti-spoofing datasets are CASIA-SURF\cite{zhang2019dataset} and CASIA-SURF CeFA\cite{li2020casia}. However, these datasets can not meet the demand for model evaluation under surveillance scenarios. Under surveillance scenarios, the images are usually low-resolution and blurry and the trained model should be deployed across different camera devices. The training set and the testing set in these datasets are in the same domain. The evaluation results on these datasets can not reflect model performance when considering the cross-device domain problem. Moreover, the data preprocessing method in these datasets involves using PRNet\cite{feng2018joint} which is time-consuming in the deployment scenario. And their preprocessing methods will cause discontinuity and can not remove depth holes caused by camera noise.

To improve the FAS performance towards low-quality images, we propose an attention-based face anti-spoofing network with feature augment (AFA). It consists of the depthwise separable attention module (DAM) and the multi-modal based feature augment module (MFAM). To further research the cross-device domain problem under surveillance scenarios, we propose a dataset called GREAT-FASD-S. We capture data using two cameras (Intel RealSense SR300 and PICO DCAM710). We propose two algorithms for processing original depth images, which can fix depth holes, normalize the depth map reasonably only based on the efficient face detector\cite{zhang2016joint}. The experiments show that our proposed AFA can achieve state-of-the-art performance compared with other methods.

To sum up, this paper makes the following contributions. 
\begin{itemize}
\item We propose an attention-based face anti-spoofing network with feature augment (AFA) which consists of the depthwise separable attention module (DAM) and the multi-modal based feature augment module (MFAM). It can boost the face anti-spoofing performance towards low-quality and cross-device domain data.
\item To research the cross-device domain problems in face anti-spoofing tasks, we propose a cross-device domain face anti-spoofing dataset called GREAT-FASD-S. Two multi-modal cameras with different imaging principles are included.
\item We propose two depth map preprocessing and normalization methods. Our method can recover the discontinuity (grid effect) and depth holes caused by cameras which are commonly found in the previous face anti-spoofing datasets. Different from previous multi-modal face anti-spoofing datasets, our preprocessing methods only use the face detector to ensure efficiency.
\item Extensive experiments demonstrate that our method can achieve state-of-the-art performance compared to other methods on the CASIA-SURF dataset and our proposed cross-device domain GREAT-FASD-S dataset.
\end{itemize}

The rest of this paper is organized as follows. Section~\ref{sect:rela-work} discusses the related work. Section~\ref{sect:dataset} introduces the cross-device domain GREAT-FASD-S dataset. Section~\ref{sect:framework} introduces the proposed face anti-spoofing framework. Section~\ref{sect:exp} provides experiments of the proposed method. Finally, Section~\ref{sect:conc} concludes this paper.

\Figure[t][width=3 in]{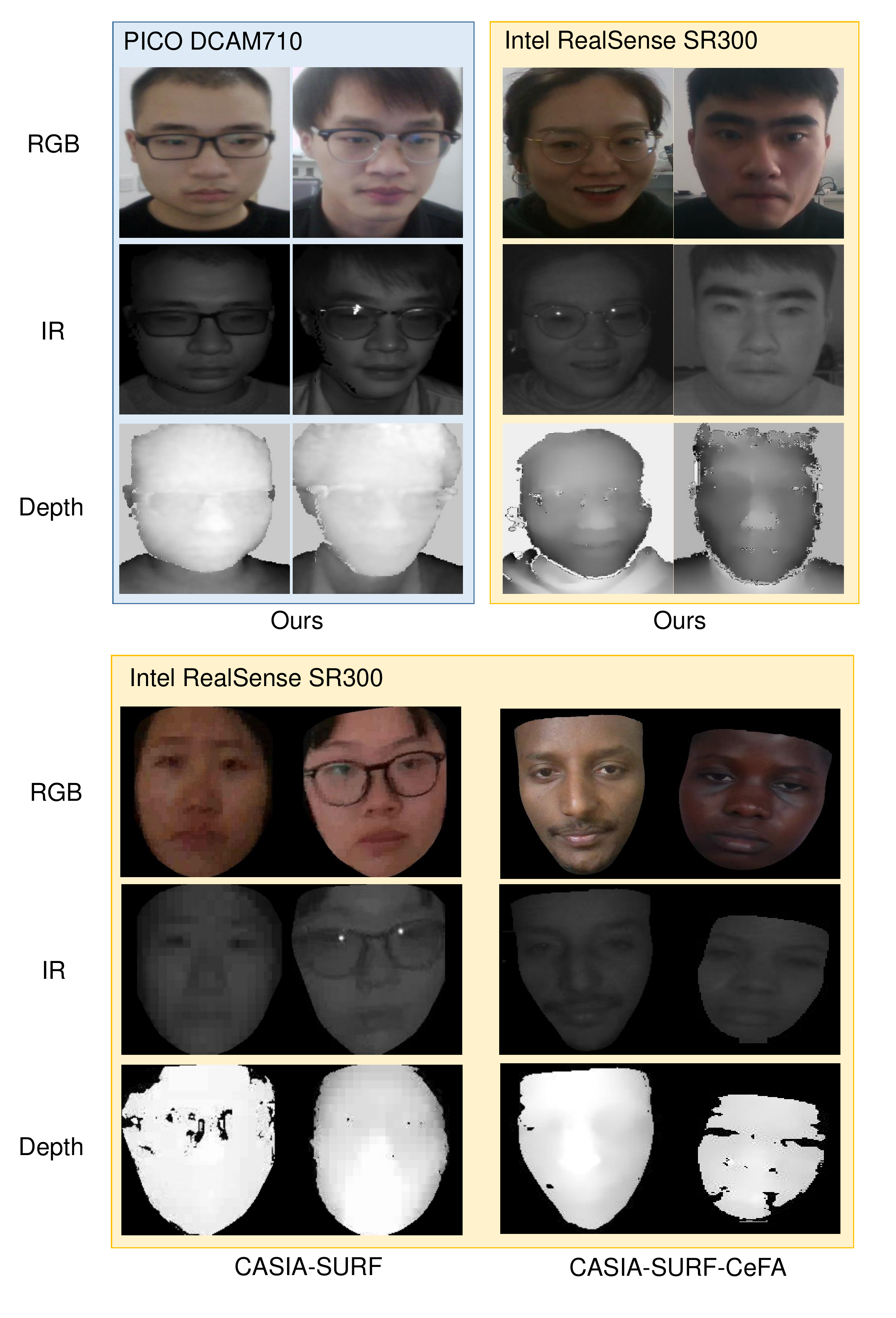}
{Visualization of different multi-modal datasets and our cross-device domain face anti-spoofing dataset.\label{fig:data_compare}}

\section{Related Work} \label{sect:rela-work}
\subsection{Single modal anti-spoofing datasets}
The mainly public available face anti-spoofing datasets are single-modal, which only contain the colorful RGB modality, including Replay-Attack\cite{chingovska2012effectiveness}, CASIA-FASD\cite{zhang2012face}, CASIA-MFSD\cite{zhang2012face} and SiW\cite{liu2018learning}. There are also some single-modal RGB datasets contain the attack method of the facial video playback using the smartphone, such as MSU-MFSD\cite{wen2015face}, Replay-Mobile\cite{costa2016replay} and OULU-NPU\cite{boulkenafet2017oulu}. 

Some of these single-modal datasets are static. They only contain discrete images. There is no correlation between pictures in the datasets. The other part is continuous. Some of these datasets directly give collected videos, some give the continuous image sequence after the frame extraction at a certain frame interval. Such data retains the correlation between neighbor frames. Researchers can use some methods based on sequence or optical flow to utilize the time domain information. Although there are many single-modal anti-spoofing datasets, the problems are obvious. We can only utilize the single modal data during face anti-spoofing tasks, which will lead to degraded performance under extreme changes in the environment and facial poses\cite{kim2015face}.
\subsection{Multi-modal anti-spoofing datasets}
Recently, there have been several works in incorporating multi-modal data to improve the performance of face anti-spoofing. CASIA-SURF\cite{zhang2019dataset} is the largest publicly available multi-modal face anti-spoofing dataset, it includes 1000 subjects with three modalities, RGB, Depth, and IR. A series of methods have achieved good face anti-spoofing performance on the CASIA-SURF dataset. 

However, this dataset has the following problems. Firstly, faces included in the dataset are all Asian faces, and the number of attack types is very small. It only includes flat paper printing and printing paper digging. Fewer attack types limit the generalization performance of the model trained with this dataset. 
Secondly, the dataset preprocessing method in this dataset involves using PRNet\cite{feng2018joint} to estimate the face area. It reconstructs the three-dimensional shape of the face and gets the face area through rendering. This process requires a lot of computing resources. In the real deployment scenario, such as surveillance scenarios, this step cannot achieve efficient face anti-spoofing during the actual application process. Besides, even if the performance of models trained on this dataset is excellent, it is also a risk of overfitting on the foreground and background separation\cite{zhang2019dataset}. If the face area is not divided during the application process, the left background noise can cause performance degradation. Thirdly, the dataset does not perform reasonable normalization preprocessing on the original face depth map, resulting in a serious grid effect on the face depth image and uneven distribution of depth data. This will lead to a consequence that the effective depth value only varies in a narrow dynamic range. 

Moreover, CASIA-SURF ignores the problem of domain adaptation, especially changes at the camera device level. Imaging principles of different multi-modal camera modules are different. For depth map, imaging principles include structured light, time of flight (ToF), \emph{etc.} Different imaging principles apply to different distance ranges. The value of the formed depth map is also different. For the IR image, the overall brightness formed by the near-infrared camera often depends on the emission power of the infrared fill light. The distance between the object and the camera will affect the intensity and quantity of infrared rays. The large infrared camera can finally capture the overall brightness of the IR image. The lack of data with changes in the device domain cannot reflect the effectiveness of face anti-spoofing algorithm that is robust to various multi-modal devices. Based on CASIA-SURF, CASIA-SURF CeFA\cite{li2020casia} captures human faces from more regions to solve the ethnic bias for face anti-spoofing. However, problems such as slow preprocessing speed, small attack types, and using a single collection device still exist. 

Compared to CASIA-SURF, we propose a cross-device domain face anti-spoofing dataset called GREAT-FASD-S. Two multi-modal cameras with different depth sensor principles are included (Intel RealSense SR300 and PICO DCAM710) to capture images. It contains human faces from 4 regions, 4 attack types including flat and 3D spoofing methods and the age range is from the 20s to 50s.
\subsection{Face anti-spoofing methods}
The deep learning method has improved the performance of face recognition, face detection, and 3D face reconstruction significantly. Compared with other biometric technologies, face anti-spoofing still has many limitations, especially in surveillance scenarios where images are usually low-resolution and blurry. 

Zhu \emph{et al.}\cite{zhu2020plenoptic} present a passive presentation attack detection method based on a complete plenoptic imaging system. They train a CNN architecture to decouple the light-field image and interference from a raw image. Then they extract hand-crafted features and train SVM classifiers to get classification results. A decision-level fusion method is also applied to SVM classifiers to further boost the performance. Li \emph{et al.}\cite{li20203d} extract the intensity distribution histograms to represent the intensity differences between the real face and 3D face mask. The 1D convolutional network is further used to capture the information. These methods are explainable and have high efficiency. However, they are not robust to environmental variance because of using hand-crafted features, such as intensity histograms which will lose the spatial information.

Liu \emph{et al.}\cite{liu2018learning} design a novel network architecture to leverage the depth map and r-PPG signal as supervision. However, the assumptions about attack methods in Liu \emph{et al.}\cite{liu2018learning} limit their generalization performance. Feng \emph{et al.}\cite{feng2020learning} reformulate FAS in an anomaly detection perspective and propose a residual-learning framework to learn the discriminative live-spoof differences which are defined as the spoof cues. Although they can achieve better generalization performance on different attack types, their method will fail under extreme environmental changes due to the single-modal data. And the spoofing cues such as r-PPG can not be detected accurately when input images are low-resolution and blurry.

Wang \emph{et al.}\cite{Wang2017RobustFA} propose a robust representation jointly modeling 2D textual information and depth information for face anti-spoofing. They use five convolutional layers and three fully-connected layers for texture feature learning and use the LBP as the depth feature. However, the hand-crafted feature limits the representation power of depth information. Their fusion scheme for multi-modal features makes it difficult to simultaneously utilize complementary information among multiple modalities in the learning process. Zhang \emph{et al.}\cite{zhang2019dataset} present a new multi-modal fusion method based on SE-block\cite{hu2018squeeze}. Zhang \emph{et al.}\cite{zhang2019feathernets} propose the FeatherNet, which improves the performance and reduces the complexity. However, FeatherNet utilizes the multi-modal data in a cascade way. It hinders the network learning features from multi modalities simultaneously. Parkin \emph{et al.}\cite{parkin2019recognizing} propose to fine-tune the model from face recognition task and use aggregation blocks to fuse the information of different modalities. Shen \emph{et al.}\cite{shen2019facebagnet} propose to solve the face anti-spoofing based on the patch-based features. Although these methods can deal with the face anti-spoofing problems well on the multi-modal CASIA-SURF dataset, the performance will drop dramatically under the surveillance scenarios where the face images are low-resolution and blurry. The performance of the above methods will further degrade when considering the cross-domain problem.

To solve the above problems, we propose the AFA which consists of the depthwise separable attention module (DAM) and the multi-modal based feature augment module (MFAM). Our method has better FAS performance under the low-resolution and blurry images and has better generalization performance across different imaging devices. 
\section{GREAT-FASD-S face anti-spoofing dataset} 
\label{sect:dataset}
\subsection{Overall}
In this section, we introduce our proposed GREAT-FASD-S dataset. We simultaneously use Intel RealSense SR300 and PICO DCAM710 cameras to capture the multi-modal videos, including RGB, depth, and infrared (IR). The depth images are processed using depth normalization algorithms. We use official alignment tools to align three modal videos. During video recording, the frame rate of both cameras is set to 30 frames per second. In the decimation process, we reserve one frame every 6 frames to make sure that the two consecutive images have certain changes. Because these multi-modal cameras have a limitation of working radius, we preprocess the collected data similar to methods\cite{dong2015image}\cite{kim2016accurate} to generate the low-quality images. As described in\cite{wheeler2011face}, a face would be imaged with a resolution of about 13 by 7 pixels under surveillance scenarios. The cropped face regions are resized to the resolution of 16x16. We also use the Gaussian kernel\cite{gaussianblur} to generate the noise appearing in surveillance cameras. Following we will introduce the process of generating the dataset in detail.
\begin{table*}
\renewcommand\arraystretch{1.2}
\centering
\caption{The effect of low resolution and blur on the CASIA-SURF dataset. The higher the value of PSNR and SSIM, the smaller the distortion. The PSNR and SSIM are lower with the decrease of image quality.}
\setlength{\tabcolsep}{22pt}
\small{
\begin{tabular}{|c|c|c|c|c|c|c|}
\hline
\multirow{2}{*}{Resolution} & \multicolumn{3}{c|}{PSNR} & \multicolumn{3}{c|}{SSIM} \\ \cline{2-7} 
                            & RGB     & DEPTH  & IR     & RGB     & DEPTH  & IR     \\ \hline \hline
64x64                       & 33.78   & 27.11  & 38.02  & 0.9876  & 0.9841 & 0.9891 \\ \hline
64x64 blur                  & 29.86   & 23.24  & 34.78  & 0.9690  & 0.9597 & 0.9759 \\ \hline
32x32                       & 29.66   & 23.05  & 34.55  & 0.9675  & 0.9578 & 0.9746 \\ \hline
32x32 blur                  & 26.07   & 19.97  & 31.47  & 0.9195  & 0.9052 & 0.9423 \\ \hline
16x16                       & 25.37   & 19.21  & 30.80  & 0.9105  & 0.8962 & 0.9350 \\ \hline
16x16 blur                  & 22.12   & 16.66  & 27.62  & 0.8083  & 0.8027 & 0.8563 \\ \hline
8x8                         & 21.98   & 16.37  & 27.47  & 0.8193  & 0.8119 & 0.8648 \\ \hline
8x8 blur                    & 18.47   & 12.97  & 23.81  & 0.6619  & 0.6821 & 0.7278 \\ \hline
\end{tabular}}
\label{tab:psnr_ssim}
\end{table*}
\subsection{The impact of low-quality images}
As shown in Table \ref{tab:psnr_ssim}, we show the image quality changes due to the resolution and blur. Similar to many image super-resolution works, we show the peak signal-to-noise ratio (PSNR) and the structural similarity index (SSIM)\cite{hore2010image} between the captured images and the 112x112 clear images. The PSNR calculates the error between corresponding pixels. The SSIM measures image similarity from brightness, contrast, and structure. Because of the limitations of the hardware, we will lose information inevitably. The PSNR and SSIM decrease with the effect of blur and lower resolution. The relative movement and loss of focus will cause blur. The low resolution and limited depth accuracy only allow us to capture part of the information of the face. The variance of light will also introduce instability. In the process of making the dataset and designing the model, we consider issues caused by blur and low resolution and try to ease these problems.
\subsection{Equipments selection}
As shown in Table \ref{tab:imaging}, we choose these two multi-modal cameras because they have different imaging principles. For Intel RealSense SR300, it uses the coded light module to capture depth images. For PICO DCAM710, it uses the time of flight (ToF) module to capture depth images. Different imaging principles lead to differences in captured depth images. Performance differences between ToF and structured light are shown in Table \ref{tab:device_diff}. The captured IR images are also different due to differences in wavelength, emission power, \emph{etc.} We compare the captured images using DCAM710 and SR300 in Figure \ref{fig:camera_compare}. We can utilize these differences to study model performance across different devices.
\begin{table}[ht]
\renewcommand\arraystretch{1.2}
\centering
\caption{Differences in the imaging principle between Intel RealSense SR300 and PICO DCAM710.}
\setlength{\tabcolsep}{11.4pt}
\small{
\begin{tabular}{|c|c|c|}
\hline
Mode  & SR300                     & DCAM710 \\ \hline \hline
Depth & Coded Light               & ToF \\ \hline
IR    & \multicolumn{2}{c|}{Differences in emission power, \emph{etc.}} \\ \hline
\end{tabular}}
\label{tab:imaging}
\end{table}
\begin{table}[ht]
\renewcommand\arraystretch{1.2}
\centering
\caption{Differences between structured light and ToF.}
\setlength{\tabcolsep}{11.4pt}
\small{
\begin{tabular}{|c|c|c|}
\hline
Attributes             & Structured Light & ToF    \\ \hline \hline
Precision              & high        & low    \\ \hline
Resolution             & high        & low    \\ \hline
Robustness to lighting & low         & high   \\ \hline
\end{tabular}}
\label{tab:device_diff}
\end{table}

\subsection{Acquisition mechanism}
In Figure \ref{fig:collecting} we show our data acquisition mechanism. In the image acquisition process, we require that the volunteer is directly in front of the camera. During collection, volunteers need to move from 0.4 meters to 1.2 meters in front of cameras, which is the distance between volunteers and cameras. Volunteers also need to make a swing larger than 45 degrees on one side in three heading directions (pitch, yaw, and roll). All volunteers should be captured under two different lighting conditions at least once. We set 4 attack methods when collecting data, including flat and 3D spoofing types. For the flat spoofing type, we use color printing, black and white printing, and electronic screen. And for the 3D spoofing type, we use the 3D paper mask. We print the color and black and white attack photos using HP LaserJet Enterprise 700 color MFP M775 printer. The printing type is laser printing. The printing paper used here is 80gsm copy paper. The product used for electronic screen attacks here is the computer screen. Examples of fake data are shown in Figure \ref{fig:fake}.
\Figure[ht][width=3.3 in]{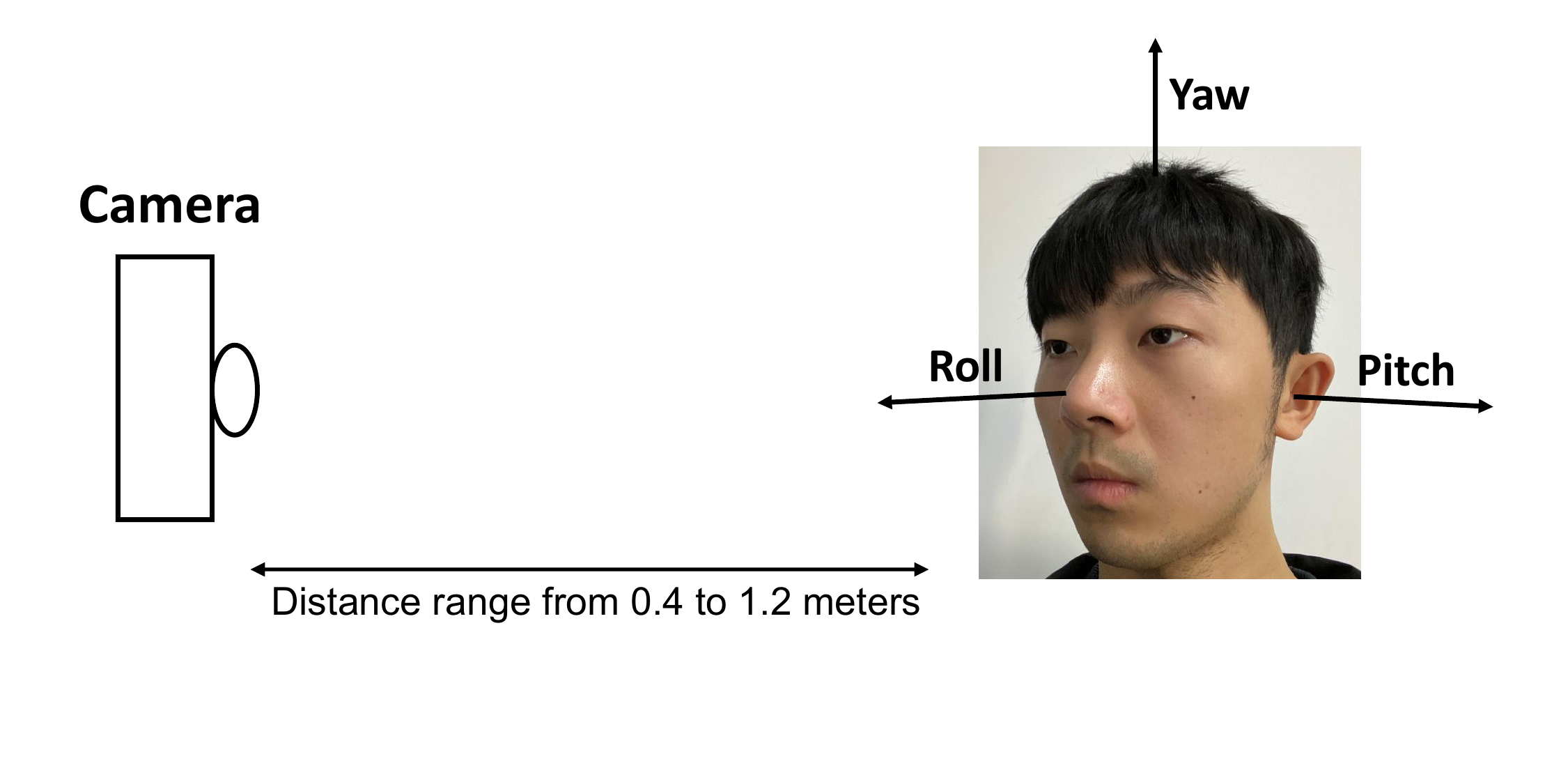}
{The diagram shows the data acquisition mechanism of our method.\label{fig:collecting}}
\Figure[ht][width=3.0 in]{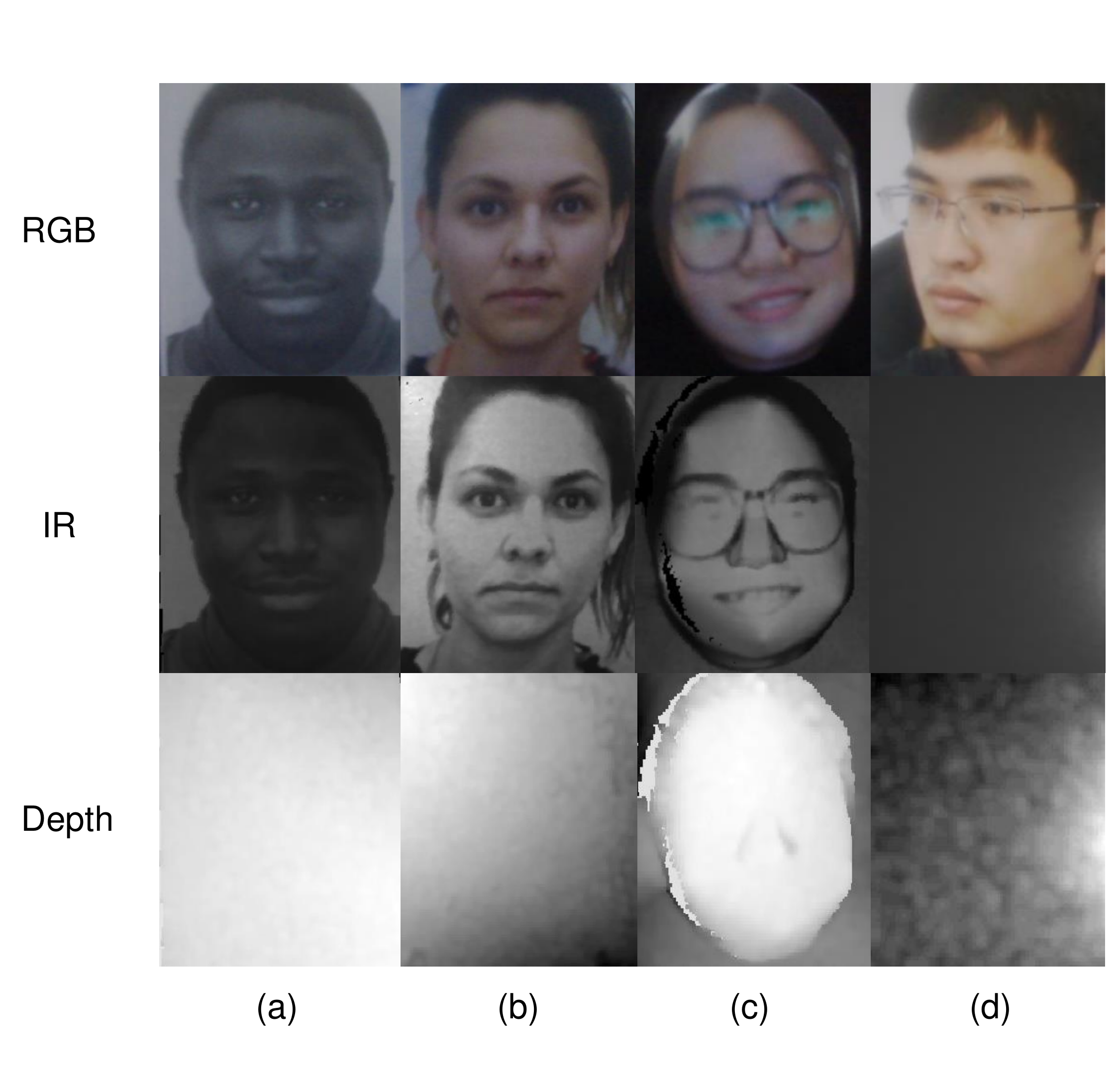}
{Examples of fake images. (a) black and white printing (b) color printing (c) 3D paper mask (d) electronic screen\label{fig:fake}}
\Figure[t][width=3.3 in]{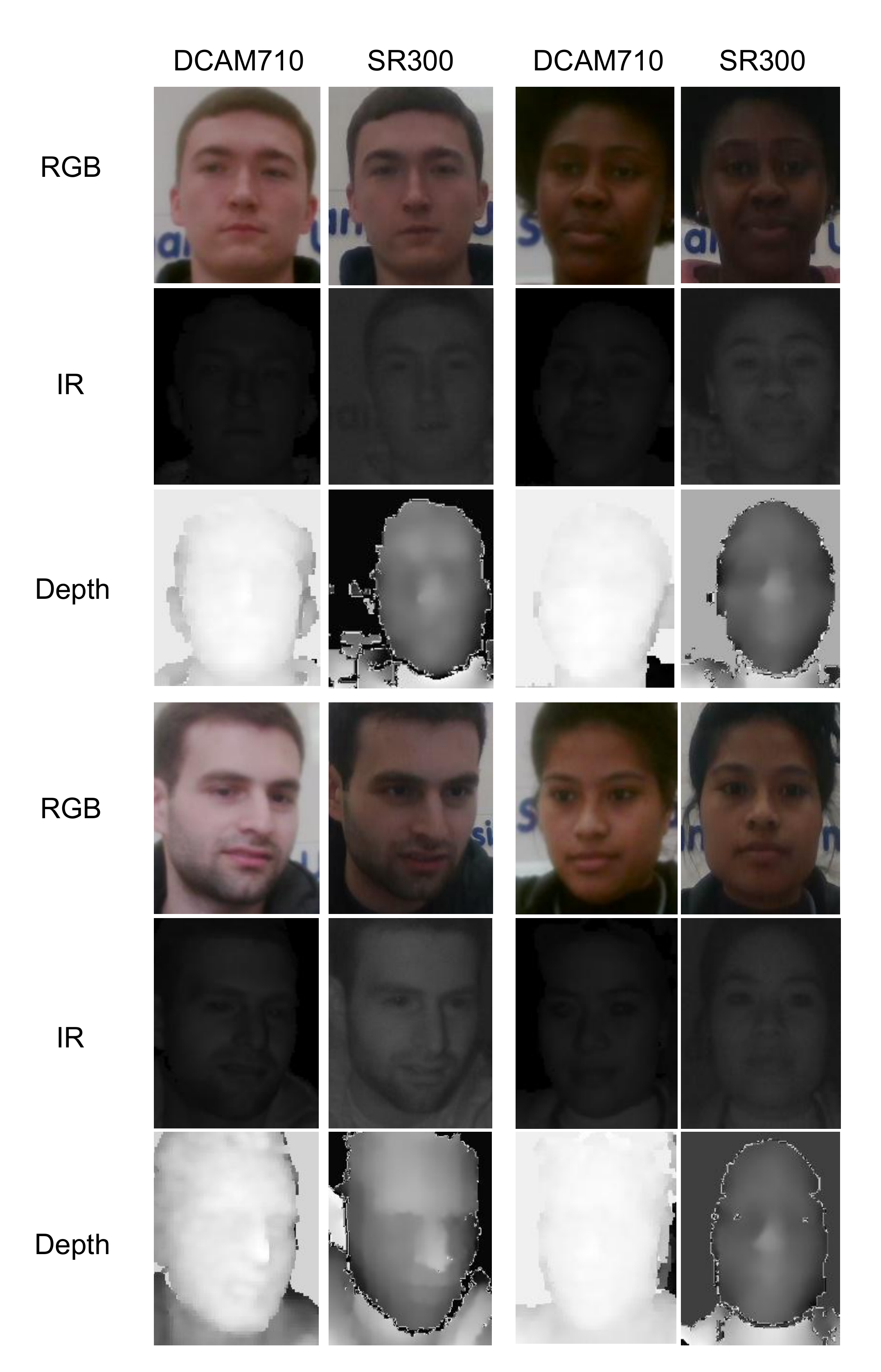}
{Visualization of multi-modal data captured using Intel RealSense SR300 and PICO DCAM710.\label{fig:camera_compare}}
\subsection{Effective area cropping procedure}
We use the MTCNN face detector\cite{zhang2016joint} here to predict the bounding box of the face area. The predicted bounding boxes of general face detectors only include the center area of faces and do not include the edge area of the face such as the forehead. Here we set an easy experiment to explore the relationship between the face anti-spoofing classification accuracy and the reasonable face region. We train and test the ResNet18-SE\cite{zhang2019dataset} on the GREAT-FASD-S dataset. The dataset contains flat and 3D spoofing types as mentioned above. Models are trained on every dataset cropped using different extension ratios. As shown in Table \ref{tab:crop}, we give the face anti-spoofing classification accuracy on every cropped set using different extension ratios. We can observe that we achieve 1.21\% higher face anti-spoofing performance using a 1.3 times expansion strategy compared to the method with no expansion. We prove that the input of face anti-spoofing networks should not be limited to the center area of faces. The whole face region and appropriate background area should be included in the input images. Therefore, we crop the origin face images based on the 1.3 times expansion region of the center area of faces. The process is shown in Figure \ref{fig:crop}.
\begin{table}[ht]
\renewcommand\arraystretch{1.2}
\centering
\caption{The effect of face region used for face anti-spoofing. The face region which is 1.3 times larger than the original output of the face detector can achieve the best performance.}
\setlength{\tabcolsep}{11.4pt}
\small{
\begin{tabular}{|c|c|}
\hline
Extension ratio & Accuracy (\%) \\ \hline \hline
1.0 & 96.87 \\ \hline
1.1 & 97.14 \\ \hline
1.2 & 97.88 \\ \hline
\bf{1.3} & \bf{98.08} \\ \hline
1.4 & 97.72 \\ \hline
1.5 & 97.69 \\ \hline
\end{tabular}}
\label{tab:crop}
\end{table}
\Figure[ht][width=3.3 in]{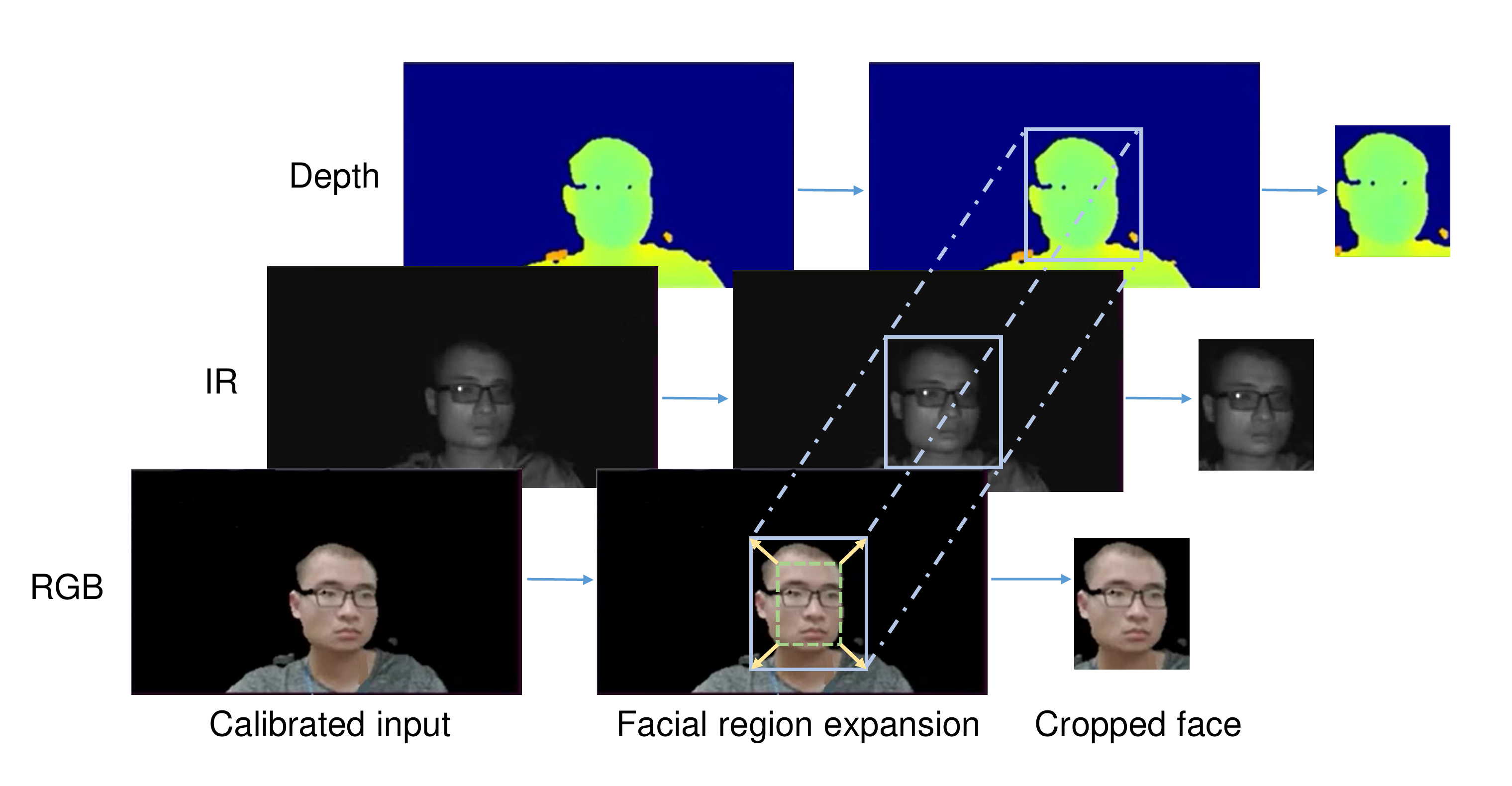}
{This figure explains the effective area cropping procedure. The bounding box generated by the face detector is widened by an extension factor, the portrait area is cropped out from the aligned three modal images. For better visualization, the depth map is mapped in color.\label{fig:crop}}
\subsection{Depth normalization algorithm}
Unlike RGB images which are usually represented using 8-bit data, depth images and IR images are generally 24 bit. To better match the current data reading interface and compare it with other methods fairly, we convert the 24-bit depth data and IR data to the 8-bit data similar to previous multi-modal datasets\cite{zhang2019dataset,li2020casia}. Because the depth data of the face area are within a small dynamic range compared to the whole 24-bit data range, we will lose the subtle difference if we use uniform quantization. Here we propose Algorithm.\ref{alg1} and Algorithm.\ref{alg2}. Our non-uniform quantization has a smaller quantization interval in the face area and a larger quantization interval in the other area. Through this way, we can keep more facial details. We can also eliminate some noise using a larger quantization interval in the non-face area. As shown in Figure \ref{fig:data_compare}, our depth images contain more facial details compared to previous datasets. For IR data, the distribution of data is relatively uniform across the whole data range. Following other multi-modal datasets\cite{zhang2019dataset,li2020casia}, we take uniform quantization to convert the IR data.

During the data collection process of the depth camera, it records the absolute distance between the target and the camera. However, for the face anti-spoofing task, we are mainly concerned about the relative depth distribution of the face region. In CASIA-SURF, they use the 3D face reconstruction method PRNet\cite{feng2018joint} to get the accurate face region during depth preprocessing. Because these datasets only publish the processed data instead of the raw images. To achieve images similar to the training data in the deployment scenario, we also need to use PRNet to preprocess the input images. This process requires a lot of computing resources. In the real deployment scenario, this step cannot achieve efficient face anti-spoofing. The speed of MTCNN face detector\cite{zhang2016joint} and PRNet\cite{feng2018joint} tested on NVIDIA Tesla P40 is shown in Table \ref{tab:speed}. For MTCNN, we measure the time from image reading to bounding box predicting (60ms). For PRNet, we measure the time from image reading to 3D shape prediction (87ms) and the time rendering the visible face region (5.5s). It shows that PRNet based depth images preprocessing method is very slow (87ms+5.5s) and can not satisfy the real-time face anti-spoofing need under deployment scenarios. Besides, due to the problem of camera noise, the depth map of the effective face area may have a problem of missing a few points of data, namely depth holes. As shown in Figure \ref{fig:data_compare}, CASIA-SURF and CASIA-SURF CeFA can not solve depth holes. Besides, their results also contain the grid effect, which is not what we want. 
\begin{table}[ht]
\renewcommand\arraystretch{1.2}
\centering
\caption{The speed of MTCNN\cite{zhang2016joint} and PRNet\cite{feng2018joint} on NVIDIA Tesla P40. For MTCNN, we measure the time from image reading to the output of the face bounding box. For PRNet, we separately count the time from image reading to the prediction of 3D shape and the time of rendering the visible face region.}
\setlength{\tabcolsep}{11.4pt}
\small{
\begin{tabular}{|c|c|}
\hline
Method & Speed \\ \hline \hline
MTCNN \cite{zhang2016joint} &   60ms    \\ \hline
PRNet \cite{feng2018joint}  &   87ms+5.5s    \\ \hline
\end{tabular}}
\label{tab:speed}
\end{table}

Therefore, we decide to explore some fast and reasonable depth images preprocessing algorithms to solve the above problems. The most straightforward method is to divide the original depth value by the max value as the new pixel value. In this way, it has a problem that the original absolute depth value is not evenly distributed over the entire number range. The depth value of the face region is concentrated in a narrow range. If we evenly quantify the original depth value across the whole value range, the smaller depth value difference in the face region will be ignored due to the excessive quantization interval, which is reflected in the visual effect. The depth map will present a grid effect. The original continuous depth map becomes a block with obvious boundaries. Besides, the converted depth map is darker or brighter overall due to the difference in the distance between the target and the camera. Following we propose two algorithms for processing original depth images, which can fix depth holes and normalize the original depth images reasonably only using the face detector. The algorithms are shown in Algorithm.\ref{alg1} and Algorithm.\ref{alg2}.
\begin{algorithm}[ht]
\caption{Depth Preprocessing Algorithm one}
\label{alg1}
\begin{algorithmic}[1]

\STATE get the face region in the depth map according to the bounding box as the face depth map;
\FOR{each $p_{i} \in$ NoneZeroFaceDepthMap}
\STATE record $maximum$ and $minimum$ values til now;
\STATE sum up all the pixel values;
\ENDFOR
\STATE calculate the average value as $mean$;
\STATE expand the bounding box to 1.3 times to get the human head including part of the background as the portrait depth map;
\FOR{each $p_{i} \in$ PortraitDepthMap}
\IF{$p_{i} = 0$}
\STATE  replace the pixel value with $mean$;
\ENDIF
\ENDFOR

\FOR{each $p_{i} \in$ DepthMap }
\STATE  get the normalized depth value $p_{ni}$ accroading to the $p_{i}$, $mean$, $maximum$, and $minimum$;
\ENDFOR
\end{algorithmic}
\end{algorithm}

\begin{algorithm}[ht]
\caption{Depth Preprocessing Algorithm two}
\label{alg2}
\begin{algorithmic}[1]
\STATE get the face region in the depth map according to the bounding box as the face depth map;
\FOR{each $p_{i} \in$ NoneZeroFaceDepthMap}
\STATE sum up all the pixel values; 
\ENDFOR
\STATE  calculate the average value as $mean$; 
\STATE  $maximum = mean+50$;
\STATE  $minimum = mean-50$;
\STATE expand the bounding box to 1.3 times to get the human head including part of the background as the portrait depth map;
\FOR{each $p_{i} \in$ PortraitDepthMap }
\IF{$p_{i} = 0$}
\STATE  replace the pixel value with $mean$;
\ENDIF
\ENDFOR
\FOR{each $p_{i} \in$ DepthMap }
\STATE  get the normalized depth value $p_{ni}$ accroading to the $p_{i}$, $mean$, $maximum$, and $minimum$;
\ENDFOR
\end{algorithmic}
\end{algorithm}
\subsection{Statistical distribution}
%
The GREAT-FASD-S dataset includes 96 real people. The captured face anti-spoofing dataset has the diversity in gender, region distribution, age, with / without glasses, and indoor lighting. We use two cameras (Intel RealSense SR300 and PICO DCAM710) to collect video clips of each person. Each person has at least 2 real video clips and 4 fake video clips. 

After removing the noise data, the data captured using the PICO DCAM710 camera contains 22,234 real-life images, in which 15,399 are for training and 6,835 are for testing. The ratio of real images to fake images captured using the PICO DCAM710 camera is controlled between 17:10. To enlarge the data distribution difference between two cameras for constructing a cross-domain dataset, the proportion between real images and fake images captured using the Intel RealSense SR300 camera is about 2:5. The age distribution range of the GREAT-FASD-S dataset is very wide, including 20 to 50 years old. The number of people in the age group of [20, 29] accounts for about 72$\%$ of all subjects. The region distribution mainly includes East Asian, European, African, and Middle Eastern, accounting for 66$\%$, 19$\%$, 8$\%$ and 7$\%$. The regional distribution and age distribution are shown in Figure \ref{fig:distri}.
\Figure[ht][width=3.0 in]{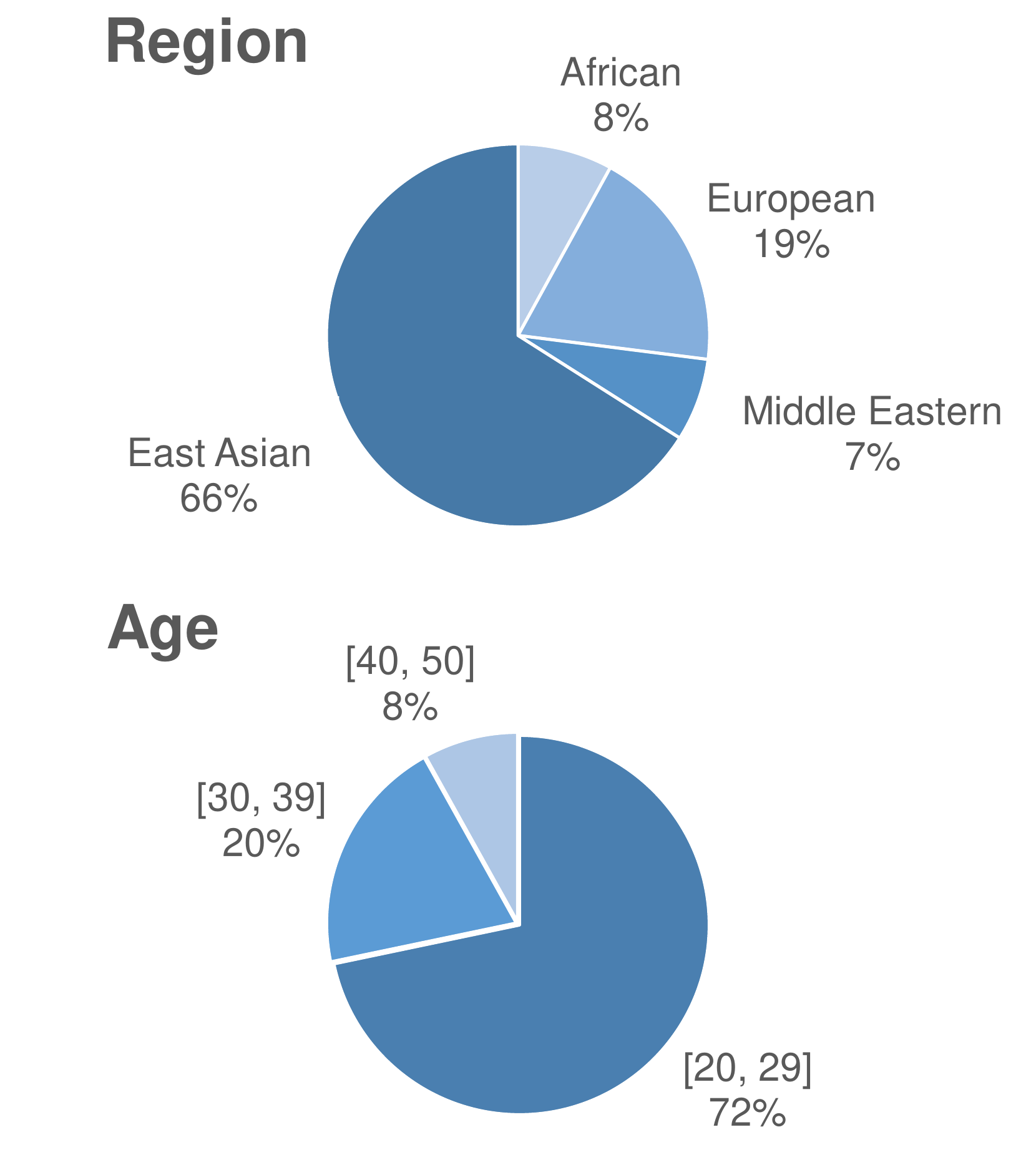}
{Region and age distribution of GREAT-FASD-S dataset.\label{fig:distri}}
\section{Proposed multi-modal face anti-spoofing Framework}
\label{sect:framework}
\Figure[ht][width=3.3 in]{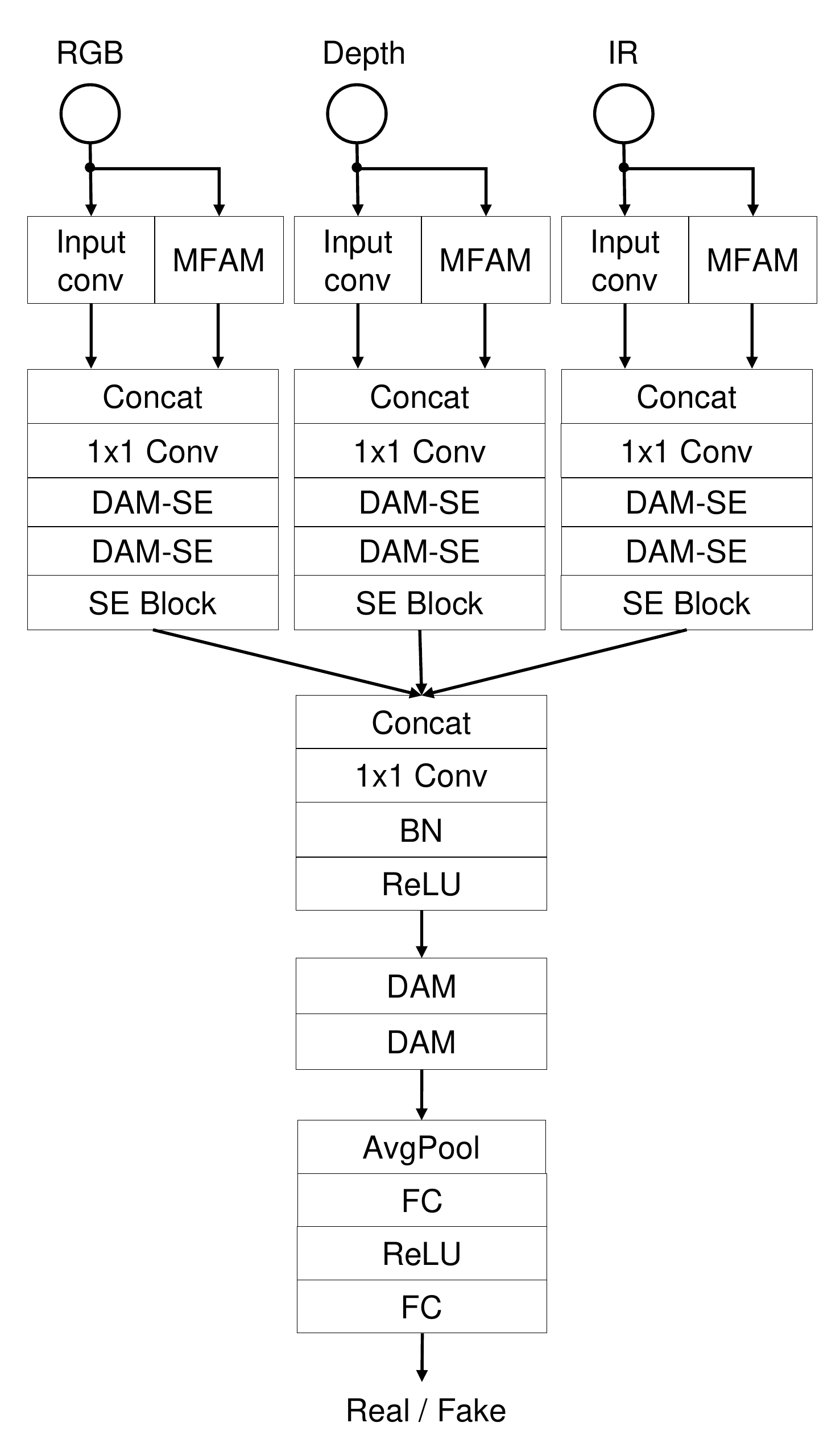}
{The network architecture of AFA. The three modal data are first processed using their own subnet. The subnet consists of MFAM and DAM-SE. The processed three modal features are re-weighted in channel dimension and concatenated together. The fused features are further processed using DAM. The global average pooling and two fully connected layers is used to get the final prediction.\label{fig:framework}}
In this section, we will demonstrate our proposed method. We will first give an overview of the proposed framework. Then we provide a detailed explanation of the depthwise separable attention module (DAM), the multi-modal based feature augment module (MFAM), and the loss function.
\subsection{Overall}
In this section, we introduce our proposed attention-based face anti-spoofing network with feature augment (AFA). The overall architecture is shown in Figure \ref{fig:framework}. Given a set of multi-modal data (RGB, Depth, IR), we first use three convolution branch to process every modal data. Every convolution branch consists of three modules: the input convolution block, the multi-modal based feature augment module (MFAM), and the depthwise separable attention module with SE-block (DAM-SE). The input convolution block consists of three identical convolution modules. Each convolution module includes one 3x3 convolution layer, one batch normalization layer, and one ReLU activation function. 

To augment the robust and representation power of extracted features, we use two sub-branches to process the input images. On the one branch, we directly extract features from origin input images. On the other branch, we use MFAM to reconstruct the features which are lost due to low-resolution and blurry. We concatenate features of two sub-branches and then use one 1x1 convolution layer to fuse features. The DAM-SE module is used to further process features. Similar to\cite{zhang2019dataset}, we use SE-block\cite{hu2018squeeze} to enhance the representational ability of different-modalities features before fusing the features of three modalities. It can explicitly model the interdependencies between different feature channels. 

After that, the fused features are processed using one 1x1 convolution layer and two DAM blocks. Based on the final convolution features, we use one global average pooling layer and two fully connected layers to get final results.
\subsection{Depthwise separable attention module}
\Figure[ht][width=2.2 in]{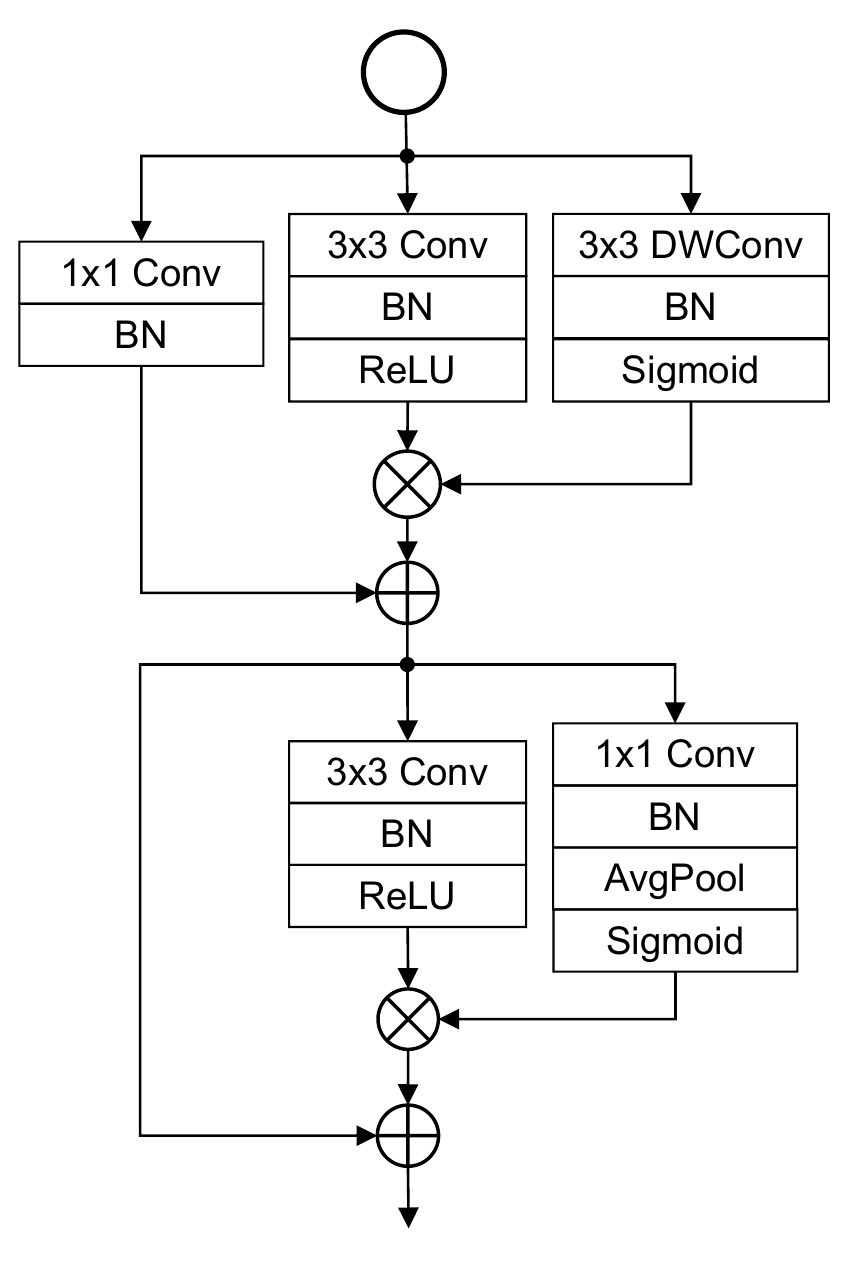}
{The depthwise separable attention module (DAM). The overall attention is separated into spatial attention and channel attention. Based on the residual block\cite{he2016deep}, the depth-wise spatial attention is added to the first block and the channel attention is added to the second block.\label{fig:dam}}
In this section, we introduce our proposed two attention-based modules, the depthwise separable attention module (DAM) and the depthwise separable attention module with SE-block (DAM-SE). There are two kinds of commonly used attention mechanism, additive attention\cite{bahdanau2014neural} and multiplicative attention\cite{vaswani2017attention}. Although the theoretical complexity of these two attention mechanisms is similar, in practice multiplicative attention is more efficient because it can be implemented using matrix multiplication. 

To avoid overfitting problems and reduce the number of parameters, inspired by\cite{howard2017mobilenets} we add the attention mechanism in a depth separability manner. The structure of DAM is shown in Figure \ref{fig:dam}. Based on the traditional residual block\cite{he2016deep}, we add two multiplicative attention branches. We first use group convolution to calculate the spatial attention map, where the group numbers equal to the greatest common divisor over the channel numbers of input and output features. Inspired by\cite{zhang2019dataset}, we can get better face anti-spoofing performance if we can balance the feature channels well. Therefore we use channel attention here to select more informative channel features. We mainly use DAM block to extract informative features from the fused multi-modal features. 

As mentioned in\cite{zhang2019dataset}, the SE-block is very useful for balancing the features of different modalities. As shown in Figure \ref{fig:dam-se}, one additional convolution layer and SE-block are added to the DAM block. To ease training, similar to residual block we add another skip connection in the DAM-SE. We use DAM-SE in every branch to further balance the features of different modalities.
\Figure[ht][width=2.6 in]{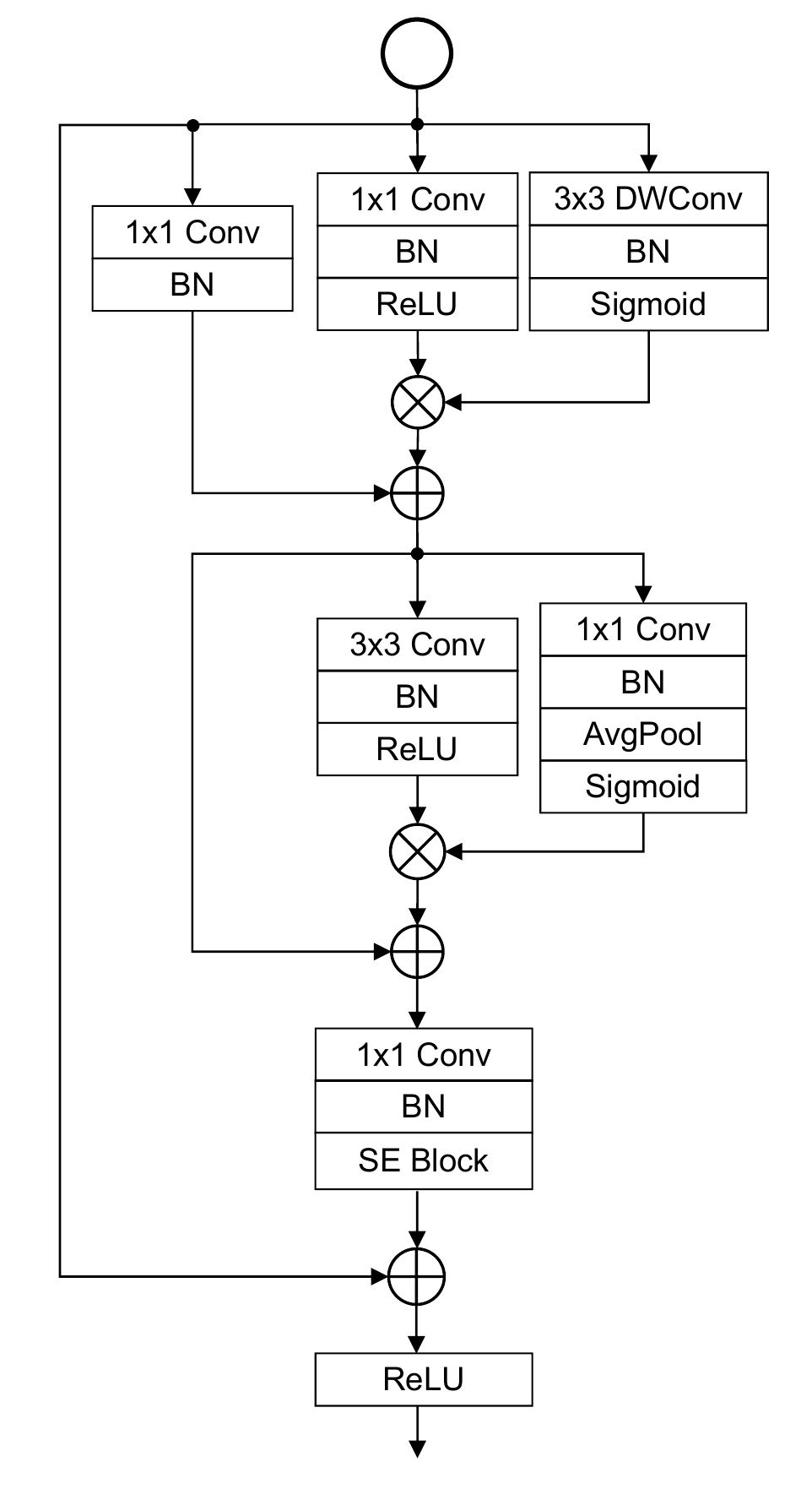}
{The depthwise separable attention module with SE-block (DAM-SE). Based on the DAM, we add the SE-block to further re-weight the different channels of features.\label{fig:dam-se}}
\subsection{Multi-modal based feature augment module}
\Figure[ht][width=2.2 in]{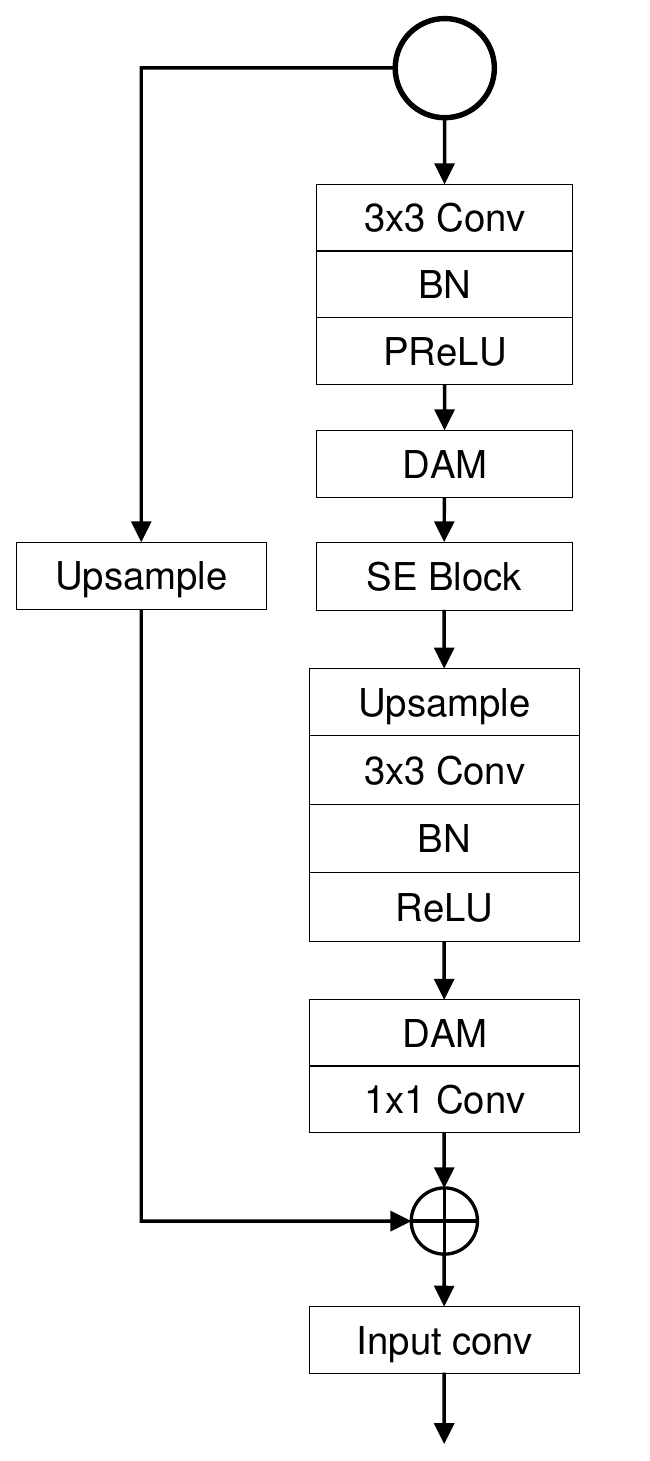}
{The single-modal based feature augment module (SFAM). The features lost due to the low-resolution and blurry are recovered based on the single-modal input images which are beneficial for the face anti-spoofing.\label{fig:sfam}}
\Figure[ht][width=2.8 in]{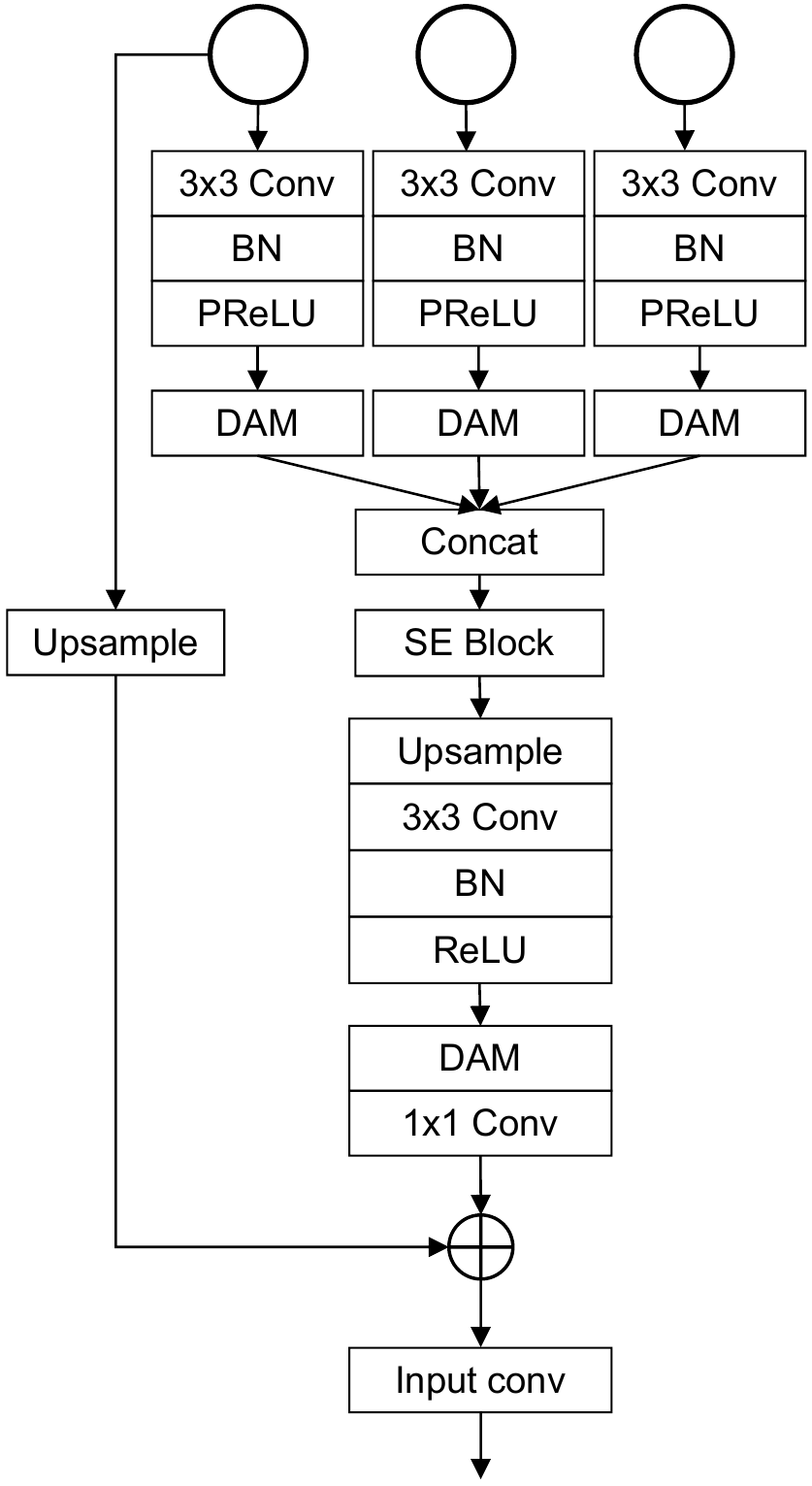}
{The multi-modal based feature augment module (MFAM). Different from SFAM, the feature is recovered based on the three modal input images to explore complementary information implied in different modal data.\label{fig:mfam}}
In this section, we introduce our multi-modal based feature augment module (MFAM). Inspired by\cite{ledig2017photo,chaitanya2019recognizing}, we hope that our network can recover discriminative features from low-resolution and blurry images which is beneficial for the final classification. To solve this problem, we first introduce a light-weight single-modal-based feature augment module (SFAM). As shown in Figure \ref{fig:sfam}, we design the SFAM to remove the noise caused by low-resolution or blur and amplify the informative features. 

Given an input image, we first use one convolution block and one DAM block to extract features from original input images. We use the SE-block to select more informative feature channels for reconstructing signals. We upsample the re-weighted features double using nearest neighbor interpolation. Then we use the other DAM block to further process the upsampled features and use one convolution layer to transform the features back to the image space. To augment the features extracted from the original input images using the input convolution block, we use the same input convolution block to map the upsampled images to feature space and concatenate these features together. To make features extracted from input images and features extracted from upsampled images have the same spatial size, the stride of input convolution block here is set to 2. 

Different from general super-resolution tasks where the input is usually one single RGB image, we have three modality data as our input in multi-modal face anti-spoofing tasks. So based on the SFAM block, we extend it to the multi-modal based feature augment module (MFAM). As shown in Figure \ref{fig:mfam}, besides using current modal data as input, we also use the other two modality data to utilize more complementary information. To select the most informative features for the current modal super-resolution task and the final classification, we use the SE-block to re-weight the feature channels of concatenated three modalities features. In experiment we show that the MFAM can further improve the face anti-spoofing performance compared to the SFAM.
\subsection{Loss}
Similar to\cite{zhang2019dataset}, we use cross entropy loss\cite{de2005tutorial} to supervise the training process of face anti-spoofing. The loss function can be described as:
\begin{equation}
\label{eqn:softmax}
\begin{split}
L_{c}&=-\sum_{i=1}^{m}log\frac{e^{W^{T}_{y_{i}}x_{i}+b_{y_{i}}}}{\sum_{j=1}^{n}e^{W^{T}_{j}x_{i}+b_{j}}}
\end{split}
\end{equation}

In order to learn robust features under the surveillance scenarios where images are usually low-resolution and blurry, we use L2 loss\cite{ng2004feature} to calculate the distance between the predicted high-resolution results of MFAM and its high-resolution groundtruth. The loss function can be described as:
\begin{equation}
\label{eqn:l2}
\begin{split}
L_{s}&=(sr_{pre}-sr_{gt})^{2}
\end{split}
\end{equation}
The final loss function can be described as:
\begin{equation}
\label{eqn:all}
\begin{split}
L&=L_{c}+\alpha L_{s}
\end{split}
\end{equation}
where $\alpha = 0.001$.

\section{Experiments} 
\label{sect:exp}
\Figure[ht][width=3.0 in]{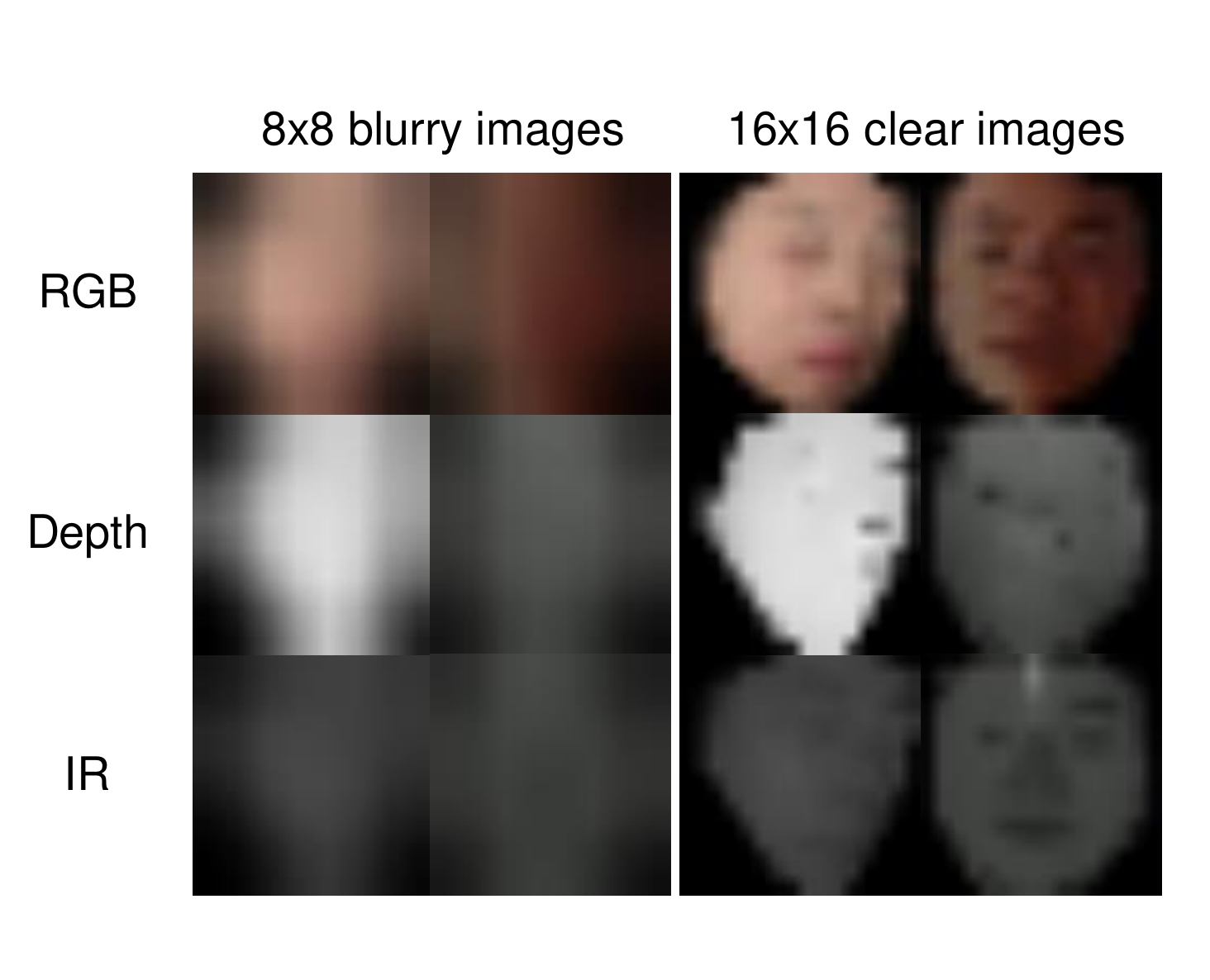}
{The samples of multi-modal data. The left images are blurry and low-resolution. The right images are clear and relatively high-resolution.\label{fig:show_lr}}
\begin{table*}
\renewcommand\arraystretch{1.2}
\centering
\caption{Results on the CASIA-SURF dataset. Models are trained on the CASIA-SURF training set and tested on the CASIA-SURF validation set.}
\setlength{\tabcolsep}{11.4pt}
\small{
\begin{tabular}{|c|c|c|c|c|c|c|}
\hline
\multirow{2}{*}{Method} & \multicolumn{3}{c|}{TPR (\%)} & \multirow{2}{*}{APCER (\%)} & \multirow{2}{*}{NPCER (\%)} & \multirow{2}{*}{ACER (\%)}\\
\cline{2-4}
 & @FPR=$10^{-2}$ &@FPR=$10^{-3}$ &@FPR=$10^{-4}$ & &  & \\
\hline
\hline
ResNet\cite{zhang2019dataset} & 88.61 & 58.82 & 32.53 & 2.59 & 3.57 & 3.08 \\
\hline
SE-Net\cite{zhang2019dataset} & 92.79 & 83.40 & 56.21 & 1.74 & 4.21 & 2.97 \\
\hline
FaceBagNet\cite{shen2019facebagnet} & 96.86 & 75.89 & 51.17 & 3.19 & 0.97 & 2.08 \\
\hline
VisionLabs\cite{parkin2019recognizing} & 94.69 & 84.84 & 61.42 & 2.27 & 1.90 & 2.09 \\
\hline
\bf{Ours} & \bf{98.56} & \bf{91.92} & \bf{81.83} & 0.92 & 1.74 & \bf{1.33}\\
\hline
\end{tabular}}
\label{tab:casia}
\end{table*}

\begin{table*}
\renewcommand\arraystretch{1.2}
\centering
\caption{Results on the GREAT-FASD-S dataset. Models are trained on the GREAT-FASD-S DCAM710 training set and tested on the GREAT-FASD-S SR300 testing set.}
\setlength{\tabcolsep}{11.4pt}
\small{
\begin{tabular}{|c|c|c|c|c|c|c|}
\hline
\multirow{2}{*}{Method} & \multicolumn{3}{c|}{TPR (\%)} & \multirow{2}{*}{APCER (\%)} & \multirow{2}{*}{NPCER (\%)} & \multirow{2}{*}{ACER (\%)}\\
\cline{2-4}
 & @FPR=$10^{-2}$ &@FPR=$10^{-3}$ &@FPR=$10^{-4}$ & &  & \\
\hline
\hline
ResNet\cite{zhang2019dataset} & 70.12 & 47.97 & 38.52 & 1.38 & 26.85 & 14.11 \\
\hline
SE-Net\cite{zhang2019dataset} & 63.32 & 42.03 & 25.99 & 3.02 & 24.15 & 13.59 \\
\hline
FaceBagNet\cite{shen2019facebagnet} & 74.50 & 46.35 & 32.63  & 4.78 & 10.43 & 7.60 \\
\hline
VisionLabs\cite{parkin2019recognizing} & 76.18 & 44.41 & 24.85 & 8.89 & 1.51 & 5.20 \\
\hline
\bf{Ours} & \bf{96.16} & \bf{85.25} & \bf{58.02} & 2.69 & 1.78 & \bf{2.24} \\
\hline
\end{tabular}}
\label{tab:great}
\end{table*}

\begin{table}[ht]
\renewcommand\arraystretch{1.2}
\centering
\caption{Comparison of weights between different models.}
\setlength{\tabcolsep}{11.4pt}
\small{
\begin{tabular}{|c|c|}
\hline
Method     & Model Weights  \\ \hline \hline
ResNet\cite{zhang2019dataset}           & 13.33$\times10^{6}$     \\ \hline
SE-Net\cite{zhang2019dataset}           & 13.34$\times10^{6}$     \\ \hline
FaceBagNet\cite{shen2019facebagnet}     & 14.55$\times10^{6}$     \\ \hline
VisionLabs\cite{parkin2019recognizing}  & 61.06$\times10^{6}$     \\ \hline
Ours                                    & 17.84$\times10^{6}$     \\ \hline
\end{tabular}}
\label{tab:model_weights}
\end{table}

\begin{table*}
\renewcommand\arraystretch{1.2}
\centering
\caption{Results on the GREAT-FASD-S dataset. Models are trained on the GREAT-FASD-S SR300 training set and tested on the GREAT-FASD-S DCAM710 testing set.}
\setlength{\tabcolsep}{11.4pt}
\small{
\begin{tabular}{|c|c|c|c|c|c|c|}
\hline
\multirow{2}{*}{Method} & \multicolumn{3}{c|}{TPR (\%)} & \multirow{2}{*}{APCER (\%)} & \multirow{2}{*}{NPCER (\%)} & \multirow{2}{*}{ACER (\%)}\\
\cline{2-4}
 & @FPR=$10^{-2}$ &@FPR=$10^{-3}$ &@FPR=$10^{-4}$ & &  & \\
\hline
\hline
ResNet\cite{zhang2019dataset} & 53.13 & 32.44 & 12.26 & 9.13 & 7.80 & 8.47 \\
\hline
SE-Net\cite{zhang2019dataset} & 81.26 & 28.56 & 3.83 & 6.69 & 4.68 & 5.69 \\
\hline
FaceBagNet\cite{shen2019facebagnet} & 80.16 & 43.25 & 9.66 & 12.53 & 4.74 & 8.63 \\
\hline
VisionLabs\cite{parkin2019recognizing} & 79.34 & 22.94 & 7.51 & 12.93 & 2.56 & 7.75 \\
\hline
\bf{Ours} & \bf{96.78} & \bf{74.57} & \bf{18.51} & 1.74 & 2.25 & \bf{1.99} \\
\hline
\end{tabular}}
\label{tab:great_transpose}
\end{table*}

\subsection{Implementation Details}
\subsubsection{Datasets}
We train and evaluate our network on two multi-modal face anti-spoofing datasets: CASIA-SURF\cite{zhang2019dataset} and our proposed GREAT-FASD-S. These datasets can demonstrate model performance from different sides. Although the CASIA-SURF only contains single-camera data, it can reflect the FAS performance within the same device domain accurately. We use CASIA-SURF to evaluate the effectiveness of the model for face anti-spoofing based on low-quality images. We use the GREAT-FASD-S to evaluate the generalization performance across different device domains.

CASIA-SURF is the largest face anti-spoofing dataset. It consists of three modalities including RGB, Depth, and IR. It contains 1,000 Chinese people in 21,000 videos. Each sample includes 1 live video clip and 6 fake video clips under different attack ways. In the different attack ways, the printed flat or curved face images will be cut eyes, nose, mouth areas, or their combinations. It removes the background except for face areas from original videos.

Our proposed GREAT-FASD-S dataset consists of three modalities including RGB, depth, and IR. We use Intel RealSense SR300 and PICO DCAM710 cameras to capture RGB, depth, and infrared (IR) video simultaneously. The age distribution range of the GREAT-FASD-S dataset is very wide, including 20 to 50 years old. The number of people in the age group of [20, 29] accounts for about 72$\%$ of all subjects. The region distribution mainly includes East Asian, European, African, and Middle Eastern, accounting for 66\%, 19\%, 8\%, and 7\%. 
\subsubsection{Experimental Settings} \label{sect:exp_settings}
The input face images are resized to the resolution of 8x8. We use gaussian kernel\cite{riba2020kornia} with $kernel\_size=3$ and $sigma=1.5$ to degrade the input images. We use random flipping, rotation, resizing, cropping and modal erasing\cite{shen2019facebagnet} for data augmentation. In the training process, we use 16x16 clear images as groundtruth in MFAM to guide our networks to learn more robust features. The training samples are shown in Figure \ref{fig:show_lr}. Our network is trained using one 2080Ti GPU with a batch size of 512. We use the Stochastic Gradient Descent (SGD) as our optimizer. The initial learning rate is set to 0.1. It decays according to the cyclic cosine annealing learning rate schedule\cite{huang2017snapshot}. Weight decay and momentum are set to 0.0005 and 0.9, respectively. We use PyTorch\cite{paszke2019pytorch} as the deep learning framework.
\subsubsection{Evaluation Metrics}
Face anti-spoofing can be considered as a classification task including two classes. Therefore, we use the accuracy of conventional classification tasks as evaluation metrics. We follow the protocols and metrics for many existing face anti-spoofing works\cite{liu2018learning, jourabloo2018face, boulkenafet2017oulu, costa2016replay}. 1) Attack Presentation Classification Error Rate (APCER), 2) Normal Presentation Classification Error Rate (NPCER), and 3) Average Classification Error Rate (ACER):
\begin{equation}
\label{eqn:APCER}
\begin{split}
APCER&=\frac{FP}{TN + FP}
\end{split}
\end{equation}
\begin{equation}
\label{eqn:NPCER}
\begin{split}
NPCER&=\frac{FN}{FN + TP}
\end{split}
\end{equation}
\begin{equation}
\label{eqn:ACER}
\begin{split}
ACER&=\frac{APCER+NPCER}{2}
\end{split}
\end{equation}
where FP, FN, TP, and TN represent false positives, false negatives, true positives, and true negatives.

Besides, similar to face recognition tasks\cite{deng2019arcface,liu2017sphereface,wang2018support}, the true positive rate (TPR) at different false positive rates (FPR) thresholds is very important in real applications. We can use the receiver operating characteristic (ROC) curve\cite{bradley1997use} to select a suitable trade-off threshold between false positive rate (FPR) and true positive rate (TPR) according to the requirements of a given real application. In the experiments, we calculate the TPR at FPR $= 10^{-2}, 10^{-3}$ and $10^{-4}$.
\begin{table*}
\renewcommand\arraystretch{1.2}
\centering
\caption{Results on the GREAT-FASD-S dataset. We present models\cite{zhang2019dataset} trained using single-modal data.}
\setlength{\tabcolsep}{11.4pt}
\small{
\begin{tabular}{|c|c|c|c|c|c|c|}
\hline
\multirow{2}{*}{Method} & \multicolumn{3}{c|}{TPR (\%)} & \multirow{2}{*}{APCER (\%)} & \multirow{2}{*}{NPCER (\%)} & \multirow{2}{*}{ACER (\%)}\\
\cline{2-4}
 & @FPR=$10^{-2}$ &@FPR=$10^{-3}$ &@FPR=$10^{-4}$ & &  & \\
\hline
\hline
RGB & 86.06 & 37.17 & 14.32 & 3.09 & 3.84 & 3.46 \\
\hline
Depth & 3.30 & 0.22 & 0.00 & 68.08 & 5.73 & 36.90 \\
\hline
IR & 6.43 & 5.19 & 4.75 & 12.18 & 85.68 & 48.93 \\
\hline
\bf{Ours} & \bf{96.16} & \bf{85.25} & \bf{58.02} & 2.69 & 1.78 & \bf{2.24} \\
\hline
\end{tabular}}
\label{tab:great_single}
\end{table*}
\begin{table*}[!ht]
\renewcommand\arraystretch{1.2}
\centering
\caption{Ablation studies on the CASIA-SURF dataset. We demonstrate the effect of our proposed MFAM and DAM. The CA means the channel attention in DAM.}
\setlength{\tabcolsep}{11.4pt}
\small{
\begin{tabular}{|l|c|c|c|c|c|c|}
\hline
\multicolumn{1}{|c|}{\multirow{2}{*}{Method}} & \multicolumn{3}{c|}{TPR (\%)} & \multirow{2}{*}{APCER (\%)} & \multirow{2}{*}{NPCER (\%)} & \multirow{2}{*}{ACER (\%)}\\
\cline{2-4}
 & @FPR=$10^{-2}$ &@FPR=$10^{-3}$ &@FPR=$10^{-4}$ & &  & \\
\hline
\hline
\bf{DAM+MFAM} & \bf{98.56} & \bf{91.92} & \bf{81.83} & 0.92 & 1.74 & \bf{1.33}\\
\hline
DAM+SFAM & 98.23 & 91.58 & 51.80 & 1.36 & 1.37 & 1.37 \\
\hline
DAM & 98.56 & 88.34 & 59.49 & 1.86 & 0.90 & 1.38 \\
\hline
\hline
\bf{MFAM+DAM} & \bf{98.56} & \bf{91.92} & \bf{81.83} & 0.92 & 1.74 & \bf{1.33}\\
\hline
MFAM+CA & 97.90 & 88.98 & 53.61 & 1.38 & 1.57 & 1.47 \\
\hline
MFAM & 97.83 & 84.60 & 51.87 & 2.21 & 0.67 & 1.44 \\
\hline
\end{tabular}}
\label{tab:ablation}
\end{table*}
\begin{table*}
\renewcommand\arraystretch{1.2}
\centering
\caption{Comparison of the performance using different upsampling operators on multiple datasets.}
\setlength{\tabcolsep}{9pt}
\small{
\begin{tabular}{|c|c|c|c|c|c|c|}
\hline
\multirow{2}{*}{Method} & \multicolumn{3}{c|}{TPR (\%)} & \multirow{2}{*}{APCER (\%)} & \multirow{2}{*}{NPCER (\%)} & \multirow{2}{*}{ACER (\%)}\\
\cline{2-4}
 & @FPR=$10^{-2}$ &@FPR=$10^{-3}$ &@FPR=$10^{-4}$ & &  & \\
\hline
\hline
\multicolumn{7}{|c|}{Training set: DCAM710, Testing set: SR300} \\
\hline
Transposed Conv\cite{zeiler2010deconvolutional} & \bf{96.87} & \bf{90.17} & \bf{68.94} & 2.07 & 2.27 & \bf{2.17} \\
\hline
Nearest Neighbor Interpolation & 96.16 & 85.25 & 58.02 & 2.69 & 1.78 & 2.24 \\
\hline
\hline
\multicolumn{7}{|c|}{Training set: SR300, Testing set: DCAM710} \\
\hline
Transposed Conv\cite{zeiler2010deconvolutional} & 95.26 & 50.56 & \bf{24.80} & 2.09 & 2.55 & 2.32 \\
\hline
Nearest Neighbor Interpolation & \bf{96.78} & \bf{74.57} & 18.51 & 1.74 & 2.25 & \bf{1.99} \\
\hline
\hline
\multicolumn{7}{|c|}{Training set: CASIA-SURF\cite{zhang2019dataset}, Testing set: CASIA-SURF\cite{zhang2019dataset}} \\
\hline
Transposed Conv\cite{zeiler2010deconvolutional} & 98.53 & 90.75 & \bf{82.00} & 2.24 & 0.87 & 1.55 \\
\hline
Nearest Neighbor Interpolation & \bf{98.56} & \bf{91.92} & 81.83 & 0.92 & 1.74 & \bf{1.33} \\
\hline
\end{tabular}}
\label{tab:upsample_transpose}
\end{table*}
\subsection{Evaluation Results}
We compare our results with ResNet18\cite{zhang2019dataset}, ResNet18-SE\cite{zhang2019dataset}, VisionLabs\cite{parkin2019recognizing} and  FaceBagNet\cite{shen2019facebagnet}. Resnet18\cite{zhang2019dataset} combines the subnetworks of different modalities at a later stage and fuses features using concatenation. Resnet18-SE\cite{zhang2019dataset} uses squeeze and excitation fusion module\cite{hu2018squeeze} to fuse each feature stream of different modalities. VisionLabs\cite{parkin2019recognizing} enriches the model with additional aggregation blocks at each feature level. Each aggregation block takes features from the corresponding residual blocks and from the previous aggregation block, making the model capable of finding inter-modal correlations not only at a fine level but also at a coarse one. FaceBagNet\cite{shen2019facebagnet} demonstrates that both patch-based feature learning and multi-stream fusion with modal feature erasing are effective methods for face anti-spoofing. We train these models on the degradation version of CASIA-SURF and GREAT-FASD-S dataset including low-resolution and blurry images. These models are trained from scratch based on the open source code provided by the author. For Facebagnet, we train and compare the model named model\_A in the author's open source code. For Visionlabs, we train and compare the single model named resnetDLAS\_A in the author's open source code. As shown in Table \ref{tab:model_weights}, we show the comparison of weights between different models.

\subsubsection{Evaluation Results on CASIA-SURF}
In this subsection, we train our model and compared methods on the degradation version of the CASIA-SURF training set, where the input images are low-resolution and blurry. The degradation method is shown in Section~\ref{sect:exp_settings}. As shown in Table \ref{tab:casia}, our method can get state-of-the-art performance on the validation set of CASIA-SURF. Compared to other methods, our method increases by a large margin on several metrics. We can improve the TPR@FPR=$10^{-4}$ from 61.42\% to 81.83\%, and improve the ACER from 2.08\% to 1.33\%. The results demonstrate that our method can learn better and robust features on the multi-modal face anti-spoofing data under the surveillance scenarios, leading to higher classification accuracy.

\subsubsection{Evaluation Results on GREAT-FASD-S}
To prove the cross-device domain capability of the model, we give two sets of experiments on the GREAT-FASD-S dataset. In the first set, we train our method and compared methods on the training set of GREAT-FASD-S which was captured using the PICO DCAM710 camera. And test the performance of models on the testing set captured using the Intel RealSense SR300 camera. In the second set, we train our method and compared methods on the training set of GREAT-FASD-S which was captured using the Intel RealSense SR300 camera. And test the performance of models on the testing set captured using the PICO DCAM710 camera. The training set and the testing set are achieved using different cameras and use different preprocessing methods. As shown in Table \ref{tab:great}, our method can achieve significant improvement in the first set of experiments. It improves the TPF@FPR=$10^{-4}$ from 38.52\% to 58.02\% and improves the ACER from 5.20\% to 2.24\%. As shown in Table \ref{tab:great_transpose}, our method can also achieve better peroformance in the second set of experiments. It improves the ACER from 5.69\% to 1.99\%. Our method has better generalization ability across different domains.

\subsection{Performance on single-modal data}
As shown in Table \ref{tab:great_single}, we show the performance of models\cite{zhang2019dataset} trained using single modal data on the GREAT-FASD-S dataset. Table \ref{tab:great_single} shows that the RGB data are most discriminative because the training data and testing data are in the same domain. Training models only using single modal data in one domain and testing in the other domain (Depth or IR) can not give reasonable results. Fusing the features of multi-modal data can boost the performance of the TPR@FPR=$10^{-4}$ significantly, which is the most important metric for the scenarios with strict false alarm rate requirements.
\subsection{Ablation Studies}
In this subsection, we discuss the effect of depthwise separable attention module (DAM) and multi-modal based feature augment module (MFAM). The results are shown in Table \ref{tab:ablation}. The experiments are separated into two groups. The first group explains the effect of the feature augment module. We compare the performance between the network with MFAM, with SFAM, and without the feature augment module. In the second group, we discuss the effect of the attention module. We show the performance of the network with DAM, with DAM only containing channel attention and without DAM. The performance of TPR@FPR=$10^{-4}$ drops from 81.83\% to 59.49\% without MFAM. And the ACER drops from 1.33\% to 1.44\% without DAM. 

We conduct more experiments to explore the impact of different upsampling operations in MFAM. In Table \ref{tab:upsample_transpose}, we show the comparison of the performance using different upsampling operators on multiple datasets. We show three sets of experiments. In the first set, models are trained on the training set of DCAM710 and tested on the testing set of SR300. In the second set, the models are trained on the training set of SR300 and tested on the testing set of DCAM710. In the third set, models are trained on the training set of CASIA-SURF and tested on the validation set of CASIA-SURF. It shows that MFAM with nearest neighbor interpolation upsampling method can achieve better performance on the metric of TPF@FPR=$10^{-2}$, TPF@FPR=$10^{-3}$ and ACER in the second and third set. The MFAM with transposed convolution can achieve better performance on the metric of TPF@FPR=$10^{-4}$ across different datasets.
\Figure[!t][width=3.0 in]{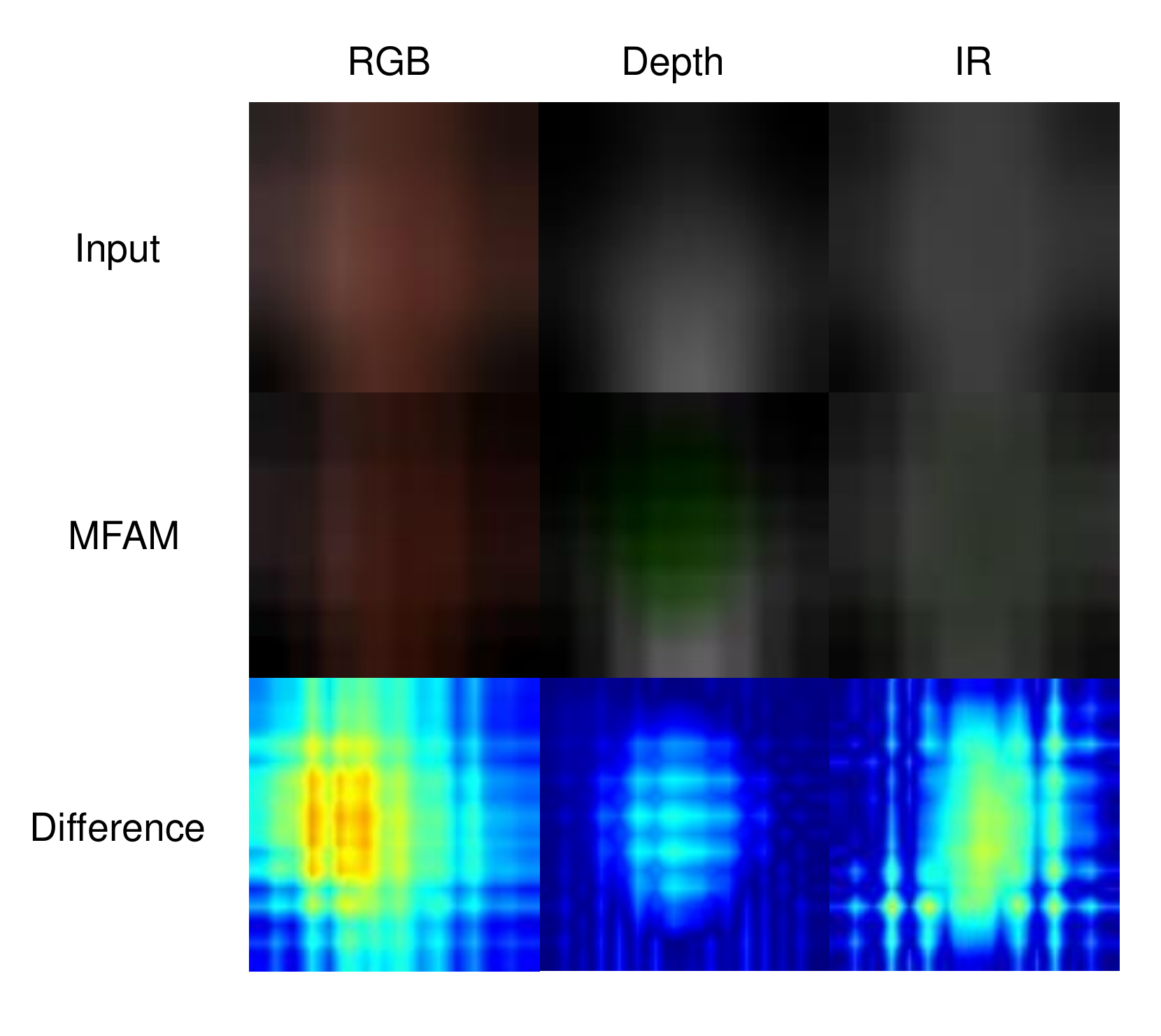}
{Visualization results. The first row is the bilinear interpolation results of the input image. The second row is the super-resolution results of the MFAM. The third row is the difference between the bilinear interpolation results and the super-resolution results of the MFAM.\label{fig:visual}}
\subsection{Visual}
In this subsection, we visual the effect of multi-modal based feature augment module (MFAM). As shown in Figure \ref{fig:visual}, we demonstrate the difference between the super-resolution result of MFAM and the bilinear interpolation\cite{kirkland2010bilinear} result of the original input image. Our MFAM can pay more attention to the foreground face region and add some high-frequency details. Although the super-resolution results of the MFAM does not look satisfactory, as shown in Table \ref{tab:ablation}, the augmented information can improve the performance of face anti-spoofing significantly.
\subsection{Discussion}
In this work, we train our multi-modal based feature augment module using paired data including clear high-resolution images and blurry low-resolution images, which will be difficult to be satisfied. Inspired by\cite{guo2020closed}, we can utilize a large amount of unpaired data to train our model in the future. We also want to introduce domain adaption methods to enhance performance. On the one hand, we can use domain adaption methods\cite{zhu2017unpaired} to transfer the source domain images to the target domain images at the pixel level. In this way, we can convert the images from different source domains to the unified target domain. On the other hand, we can use domain adaption methods\cite{ganin2015unsupervised} at the feature level. If we can get features that are invariant to the shift between the domains, we will get better cross-device domain performance. 
\section{Conclusion} \label{sect:conc}
In this paper, we propose an attention-based face anti-spoofing network with feature augment (AFA) which consists of the depthwise separable attention module (DAM) and the multi-modal based feature augment module (MFAM). The depthwise separable attention module (DAM) can select more informative features that are useful for final classification. The multi-modal based feature augment module (MFAM) utilize the complementary information implied in the three modal data to recover the high-frequency details of the current modal data. Extensive experiments prove that our method can achieve state-of-the-art performance compared to other methods. Moreover, we establish a cross-device domain multi-mode face anti-spoofing dataset called GREAT-FASD-S. It focuses on the impact of differences in hardware devices on the actual deployment of face anti-spoofing models. The dataset covers people coming from 4 regions, 4 kinds of spoofing types, wide age distribution, and complex environmental conditions. We believe this dataset will promote the development of face anti-spoofing. 

\section*{Acknowledgment}
This work was supported by the National Natural Science Foundation of China (NSFC) under Grants 62071284, 61871262, 61901251 and 61904101, the National Key Research and Development Program of China under Grants 2017YEF0121400 and 2019YFE0196600, the Innovation Program of Shanghai Municipal Science and Technology Commission under Grant 20JC1416400, and research funds from Shanghai Institute for Advanced Communication and Data Science (SICS). 

\bibliographystyle{IEEEtran}
\bibliography{access}

\begin{IEEEbiography}[{\includegraphics[width=1in,height=1.25in,clip,keepaspectratio]{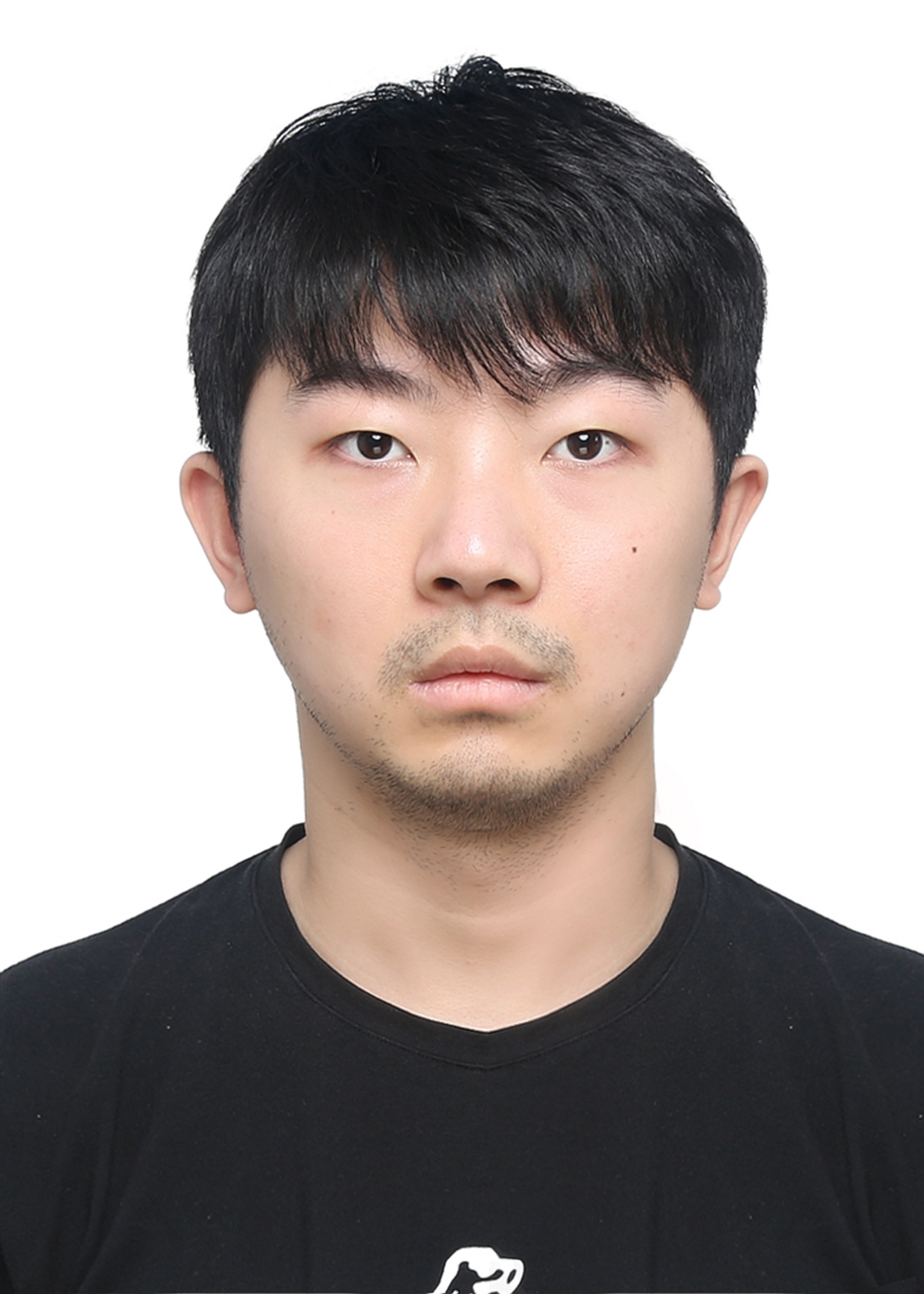}}]{Xudong Chen}
received the B.E. degree from the Department of Communication Engineering, Shanghai University, Shanghai, China, in 2018. He is currently pursuing the master degree in information and communication engineering at Shanghai University, China. His research interests include face anti-spoofing, 3D face reconstruction, face detection and face recognition.
\end{IEEEbiography}

\begin{IEEEbiography}[{\includegraphics[width=1in,height=1.25in,clip,keepaspectratio]{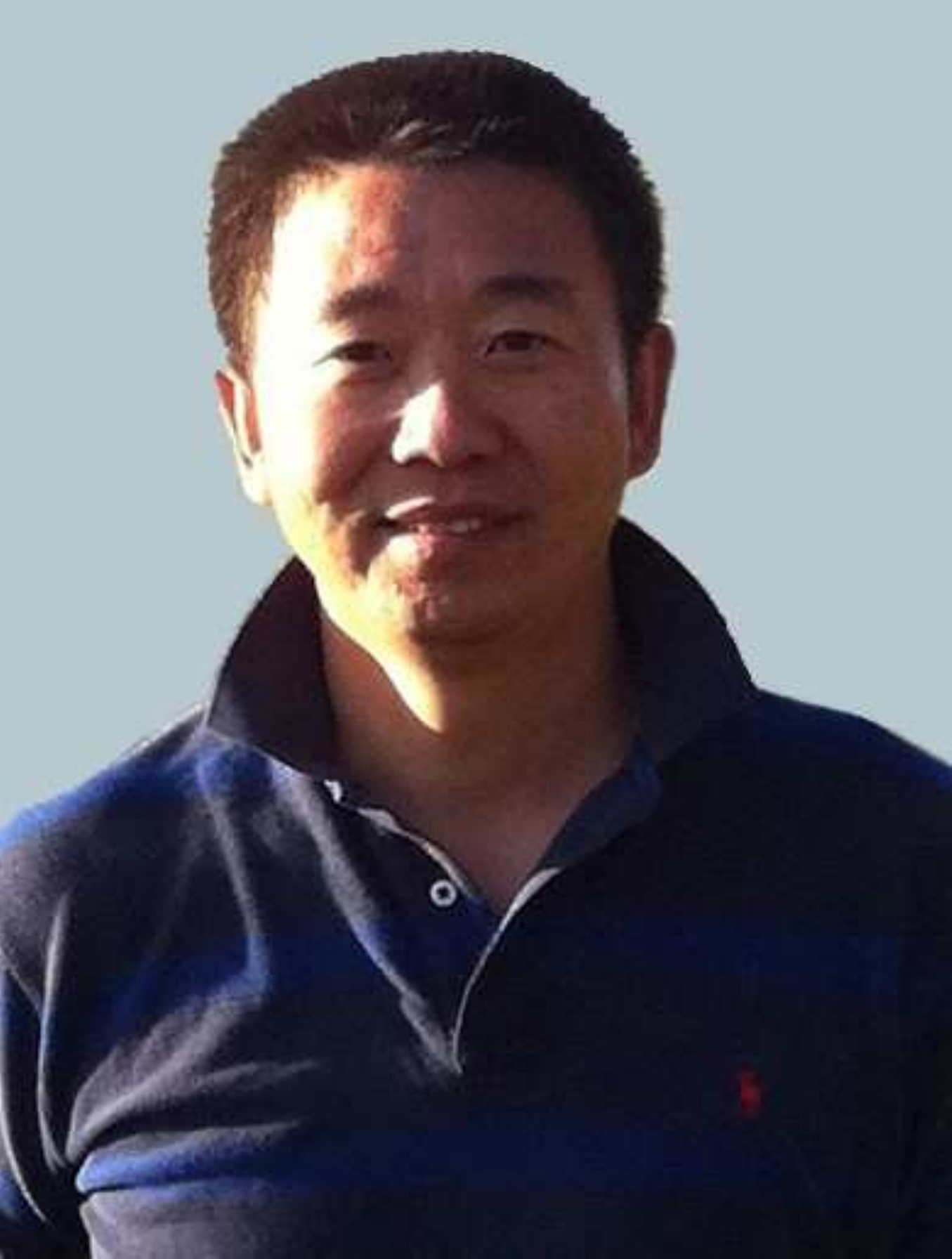}}]{Shugong Xu}
(M'98-SM'06-F'16) graduated from Wuhan University, China, in 1990, and received his Master degree in Pattern Recognition and Intelligent Control from Huazhong University of Science and Technology (HUST), China, in 1993, and Ph.D. degree in EE from HUST in 1996.He is professor at Shanghai University, head of the Shanghai Institute for Advanced Communication and Data Science (SICS). He was the center Director and Intel Principal Investigator of the Intel Collaborative Research Institute for Mobile Networking and Computing (ICRI-MNC), prior to December 2016 when he joined Shanghai University. Before joining Intel in September 2013, he was a research director and principal scientist at the Communication Technologies Laboratory, Huawei Technologies. Among his responsibilities at Huawei, he founded and directed Huawei's green radio research program, Green Radio Excellence in Architecture and Technologies (GREAT). He was also the Chief Scientist and PI for the China National 863 project on End-to-End Energy Efficient Networks. Shugong was one of the co-founders of the Green Touch consortium together with Bell Labs etc, and he served as the Co-Chair of the Technical Committee for three terms in this international consortium. Prior to joining Huawei in 2008, he was with Sharp Laboratories of America as a senior research scientist. Before that, he conducted research as research fellow in City College of New York, Michigan State University and Tsinghua University. Dr. Xu published over 100 peer-reviewed research papers in top international conferences and journals. One of his most referenced papers has over 1400 Google Scholar citations, in which the findings were among the major triggers for the research and standardization of the IEEE 802.11S. He has over 20 U.S. patents granted. Some of these technologies have been adopted in international standards including the IEEE 802.11, 3GPP LTE, and DLNA. He was awarded `National Innovation Leadership Talent' by China government in 2013, was elevated to IEEE Fellow in 2015 for contributions to the improvement of wireless networks efficiency. Shugong is also the winner of the 2017 Award for Advances in Communication from IEEE Communications Society. His current research interests include wireless communication systems and Machine Learning.
\end{IEEEbiography}

\begin{IEEEbiography}[{\includegraphics[width=1in,height=1.25in,clip,keepaspectratio]{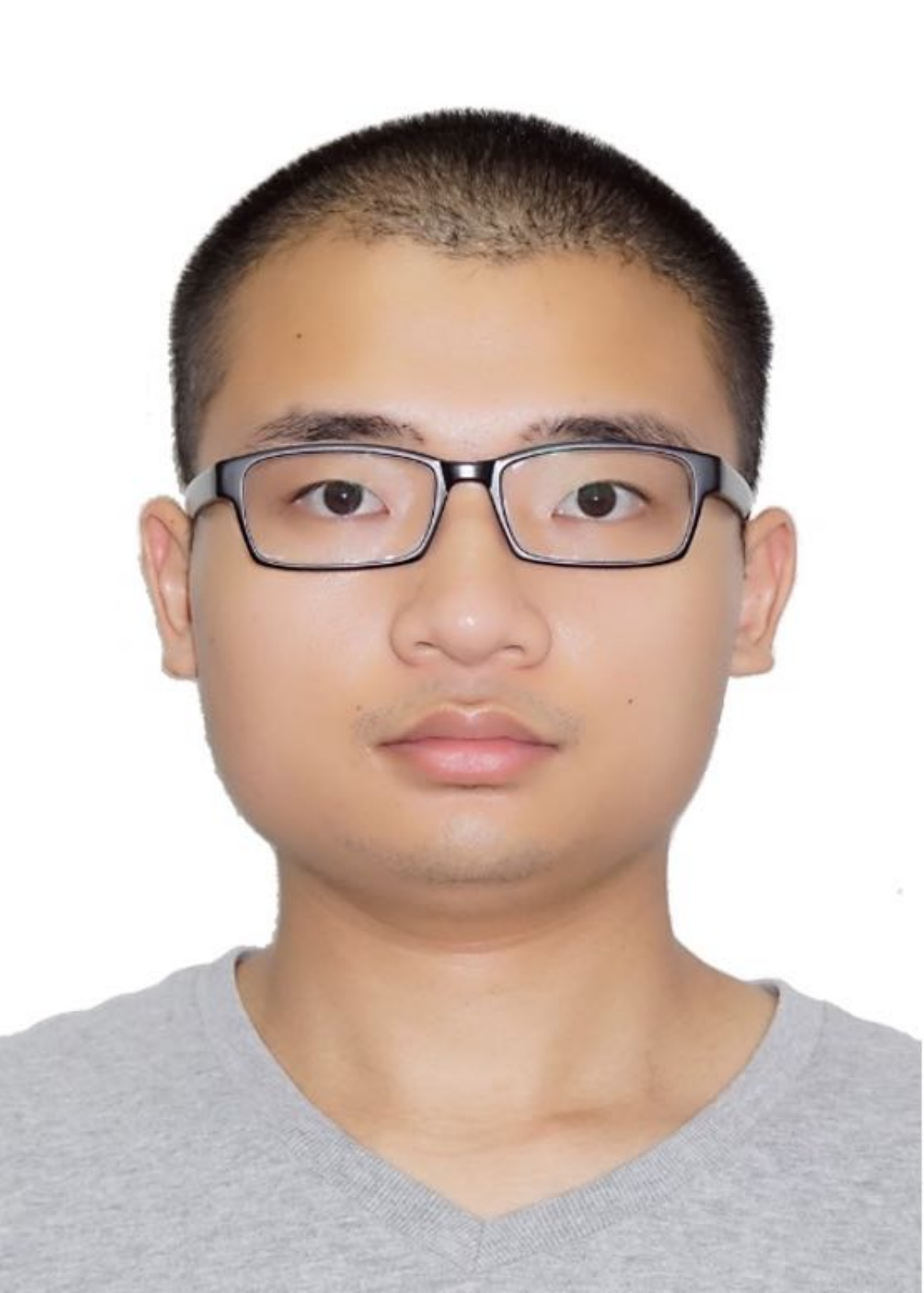}}]{Qiaobin Ji}
received the B.E. degree from the Department of Communication Engineering, Shanghai University, Shanghai, China, in 2018. He is currently pursuing the master degree in information and communication engineering at Shanghai University, China. His research interests include face recognition, face detection and face anti-spoofing.
\end{IEEEbiography}

\begin{IEEEbiography}[{\includegraphics[width=1in,height=1.25in,clip,keepaspectratio]{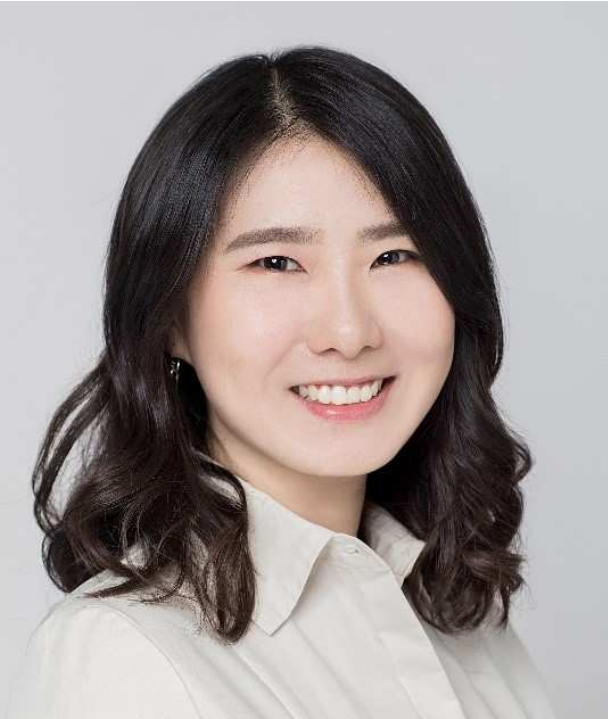}}]{Shan Cao}
received her B.S. degree and Ph.D. degree in Microelectronics from Tsinghua University, China, in 2009 and 2015 respectively. She was a postdoc in School of Information and Electronics, Beijing Institute of Technology during 2015 and 2017. She is currently an assistant professor in Shanghai University. Her current research interests include wireless communication systems, channel encoding and decoding, machine learning acceleration and its ASIC design.
\end{IEEEbiography}

\EOD
\end{document}